%% file: verdict.tex
\documentclass[11pt]{article}
\usepackage[preprint]{acl}
\usepackage{times}
\usepackage{latexsym}
\usepackage[T1]{fontenc}
\usepackage[utf8]{inputenc}
\usepackage{microtype}
\usepackage{booktabs}
\usepackage{adjustbox}
\usepackage{pgfplots}
\pgfplotsset{compat=1.17}
\usepackage{amsmath}
\usepackage{graphicx}
\usepackage{xcolor}

\title{Verdict: Encode Once, Decide Many}

\author{
  Avdhesh Singh Chouhan \\
  LEC AI \\
  avdhesh.chouhan@lecai.ai
  \And
  Jack Perry \\
  LEC AI \\
  jack@lecai.ai
}

\newcommand{\coderepo}{\url{https://github.com/LEC-AI/verdict}}

\begin{document}
\maketitle

\input{sections/abstract}

\section{Introduction}
\label{sec:intro}

Organizational automation runs on a stream of small, typed decisions made about one piece of state at a time: an email
(does it need a reply? how urgent is it? who should own it?), an invoice (pay, hold or reject? how material is the
discrepancy?), a security alert, a customer thread, an agent run. The output space of each decision is declared in
advance, and the operator needs a probability distribution over that space rather than text, so that a system can
act on its own above a confidence threshold and escalate below it \citep{geifman2017selective}. Two properties decide
whether such a decision layer is usable: the probabilities must be \emph{calibrated}, so that a threshold chosen
yesterday still means the same thing today \citep{guo2017calibration}, and the layer must be \emph{cheap}, because a
single piece of state typically triggers several questions and a deployment answers millions of them.

A recent family of ``System-1'' typed-decision models \citep{kahneman2011thinking} targets exactly this contract: the caller
declares typed questions (\emph{choice}, yes/no \emph{noul}, ordinal \emph{score}) with a written description for every
option, and receives a distribution per question from one non-generative forward pass
\citep{laya2026,jev2026,decider2026}. The open member of this family, Laya \citep{laya2026}, and the per-question
cross-encoders it resembles \citep{devlin2019bert,liu2019roberta} read the state once \emph{per question}: five questions
about one email cost five passes over the email.\footnote{Laya's documentation calls this ``one single forward pass''; we use \emph{pass} for one full encoding of
the state (Section~\ref{sec:model}), compatible if that call still allocates one encoding per question internally,
which we did not confirm from Laya's source.}

We present \textbf{Verdict}, an open decision layer built around a different computation, which we call \textbf{encode once, decide
many} (Verdict-MQ). One ModernBERT \citep{warner2024modernbert} pass reads the state and every question asked about it;
a structured attention mask makes the state encoding independent of the questions, keeps questions from influencing one
another, and gives all options of a question the same positions so that option order cannot matter. These properties
hold by construction, and we verify them numerically.

Our contributions:
\begin{enumerate}
\item \textbf{Verdict-MQ}, a packed typed-decision encoder (Section~\ref{sec:model}): one pass answers all of a case's
questions, invariant by construction to option order and question grouping (in exact arithmetic, given identical
retained state tokens), and 2.7--2.8$\times$ faster per case than one pass per question at batch size 8 (1.3--1.8$\times$
at batch size 1). We also identify a failure mode: when option markers attend to each other, they
collapse to a single representation.
\item \textbf{A distributional advantage over the open specialist} (Section~\ref{sec:typed}): on the typed-decisions
benchmark \citep{typeddecisions2026}, a 395M Verdict-MQ matches the soft labels significantly better than the released
421M Laya specialist on every fresh seed, an advantage that survives per-system temperature scaling, and a 150M model ties Laya on
accuracy. Both large systems' accuracy is at or past the benchmark's saturation point, so accuracy margins there
mainly measure agreement with its labelling teacher; held-out raw-confidence thresholds let the large model auto-handle 62--71\% of
decisions at a 90\% target against Laya's 56\% (per decision, not per case).
\item \textbf{What the shared state costs} (Sections~\ref{sec:attribution} and~\ref{sec:standard}): the accuracy lead
over Laya is most consistent with the NLI initialization, not the architecture, while a controlled $k$=0 vs.\ $k$=22
comparison shows that sharing one question-independent state costs 0.9--3.5 points on RACE and MultiRC and nothing
measurable on typed decisions, with the invariance guarantees holding at every depth.
\item \textbf{Standard benchmarks with measured references} (Section~\ref{sec:standard}): on RACE, BoolQ and MultiRC
\citep{lai2017race,clark2019boolq,khashabi2018multirc}, against a matched fine-tuned baseline Verdict-MQ wins on
MultiRC, loses narrowly on RACE and loses clearly on BoolQ, where its training is unstable. We measure zero-shot references
ourselves rather than citing vendor cards: Laya's general checkpoint, and Jev, a closed general system that beats every
fine-tuned encoder here while local inference is one to two orders of magnitude faster per case than its API.
\end{enumerate}

\section{Related Work}
\label{sec:related}

\paragraph{Typed-decision models.} Laya \citep{laya2026} (open, ModernBERT-based), Jev \citep{jev2026} and meraGPT
Decider \citep{decider2026} (closed) expose the typed-question contract we adopt. We measure Laya and Jev ourselves
(Jev through its public API on Vercel's AI Gateway, on the same items and questions as Laya: 387 of 400 test cases
succeeded, 13 lost to transient provider errors) and cite Decider's figure as reported on the benchmark card (we did
not measure it); the card also lists Featherless's Simple Jev, a zero-shot LLM-based classifier (0.716). Our measured Jev (0.734) closely matches the card maintainer's own measurement (0.727).
The typed-decisions benchmark \citep{typeddecisions2026} distinguishes \emph{specialists}, fitted on its four workflows,
from \emph{generalists} scored zero-shot, and warns that its labels come from a small teacher model.

\paragraph{Multiple choice and cross-encoders.} Scoring each option with a cross-encoder over (context, question, option)
is the standard recipe for multiple-choice reading comprehension \citep{devlin2019bert,liu2019roberta,lai2017race}; it
re-encodes the context once per option. Zero-shot classification by entailment \citep{yin2019zeroshot} and large
multi-task NLI training \citep{sileo2024tasksource} turn label descriptions into text the encoder reads. Option-order
sensitivity is documented for LLM multiple-choice selection \citep{pezeshkpour2023order,zheng2024mcq}; we find the same
failure in a fine-tuned encoder and remove it architecturally. Set-LLM \citep{egressy2025setllm} uses the two
ingredients closest to ours, SetPE (shared starting position per set element, our rule 4) and SetMask (blocked
attention between set elements, our rule 3), applied to a decoder LLM's option set for one generated multiple-choice
answer, and UniMC \citep{yang2022unimc} isolates an encoder's options with an attention mask and reads each out at its
own marker token for zero-shot classification, an encoder precedent for rules 3--4 that we had not found when
designing them. Neither packs more than one question into a pass. Our contribution is not the invariance mechanism, established
here for both decoders and encoders, but packing several \emph{independent} questions about one state into a single
pass, and measuring what that shared encoding costs and gains (Section~\ref{sec:standard}'s $k$=0/$k$=22 comparison);
a head-to-head comparison against Set-LLM or UniMC on typed decisions is future work.

\paragraph{Concurrent work.} Four preprints appeared in late September and early October 2026, as this paper was being
finalized; we do not compare against them. LAVOIR \citep{yilmaz2026lavoir} extends Laya to ask for missing
information, PACT \citep{lin2026pact} is a training objective for single-token typed decisions, and SBERT2S1
\citep{deka2026sbert2s1} converts biomedical retrieval encoders into typed-decision models; they address when to ask,
the objective and the initialization rather than the shared encoding. DuplexJev \citep{jin2026duplexjev} is the
closest to Verdict-MQ: it encodes one audio clip once and packs all of its question suffixes into one row of a frozen
causal LLM under a block-diagonal mask, reading each question out as a single-token distribution. It targets speech
with a decoder backbone, where the causal prefix gives a question-independent state for free; Verdict-MQ obtains the
same property in a bidirectional encoder through its mask, adds option-order invariance through shared option
positions (Section~\ref{sec:model}), and measures what the shared state costs (Section~\ref{sec:standard}).

\paragraph{Shared and precomputed encodings.} DeFormer \citep{cao2020deformer}, PreTTR \citep{macavaney2020prettr}, and
ColBERT/Poly-encoders \citep{khattab2020colbert,humeau2020poly} each compute part of the input independently and defer
interaction to cheap late stages. Packing independent sequences into one pass with block-diagonal masks and reset
positions is standard for efficient training \citep{krell2021packing}, batch prompting asks an LLM several questions per
call \citep{cheng2023batch}, and shared-prefix KV-cache reuse gives decoders a cached, question-independent context
\citep{zheng2023sglang,gim2024promptcache}. Verdict-MQ brings these ideas to typed decisions: the state is encoded
question-independently at every layer while several questions share one pass. Permutation-equivariant set processing
\citep{zaheer2017deepsets,lee2019settransformer} motivates the shared option positions and the optional option mixer.

\paragraph{Calibration and selective prediction.} We report ECE \citep{naeini2015ece}, Brier score \citep{brier1950},
distance to the benchmark's soft gold, and risk--coverage curves \citep{geifman2017selective}; post-hoc temperature
scaling \citep{guo2017calibration} is the standard correction we use for fair comparison. Soft-label training
\citep{peterson2019human,hinton2015distilling} and an earth-mover loss for ordinal targets \citep{hou2016emd} are our
training signals. Fine-tuning instability across seeds is documented for large encoders
\citep{dodge2020finetuning,mosbach2021stability}; intermediate-task training
\citep{phang2018stilts,pruksachatkun2020intermediate} explains why an NLI-trained starting point helps.

\section{Model}
\label{sec:model}
\input{sections/model}

\section{Experimental Setup}
\label{sec:setup}
\input{sections/setup}

\section{Results on Typed Decisions}
\label{sec:typed}
\input{sections/typed}

\section{What the Architecture Contributes}
\label{sec:attribution}
\input{sections/attribution}

\section{Standard Benchmarks}
\label{sec:standard}
\input{sections/standard}

\paragraph{Supplementary experiments.} Verdict-MQ ties standard classifiers on single-utterance intent classification
(CLINC150, Banking77; Appendix~\ref{app:standard-methods}). We also validated on public organizational email (Enron),
diagnosed a failed synthetic-corpus generalist, and logged evaluation pitfalls; none of these bears on the central
shared-state tradeoff (Appendix~\ref{app:supplementary}).

\section{Conclusion}
\input{sections/conclusion}

\section*{Limitations}
\label{sec:limitations}
\input{sections/limitations}

\section*{Ethical Considerations}
\input{sections/ethics}

\bibliography{verdict}

\appendix
\input{sections/appendix}

\end{document}

%% file: sections/abstract.tex
\begin{abstract}
Automating organizational work means answering many small typed questions about one piece of state (does this email
need a reply, how urgent is it, who owns it) with probabilities good enough to act on without a human. The open
typed-decision model we compare against, like a standard cross-encoder, answers each question with a separate pass over the input. We present \textbf{Verdict-MQ}, an
open encoder that reads a case once and answers all of its questions: given identical retained state tokens after
truncation (verified on every reported case), a structured attention mask makes the state encoding independent of the
questions, keeps questions from influencing each other, and makes option order irrelevant by construction. On the
typed-decisions benchmark, a 395M Verdict-MQ matches the soft gold labels significantly better than the released 421M
Laya specialist on all five fresh seeds (KL 0.080 vs 0.117, Brier 0.045 vs 0.061), an advantage that survives
per-system temperature scaling; its higher accuracy (0.801 vs 0.767) lies past the benchmark's own saturation point,
so we read that margin as closer agreement with the benchmark's labelling teacher. On RACE, BoolQ and MultiRC, against
a matched fine-tuned baseline it wins on MultiRC and loses on RACE and on BoolQ, where its training is unstable; we
measure zero-shot references ourselves, including a closed general system (Jev) that beats every fine-tuned encoder.
A controlled comparison measures what the shared state costs: giving each question its own question-aware copy of
the state gains 0.9--3.5 points on RACE and MultiRC at 1.4--2.5$\times$ the per-case latency, and nothing measurable
on typed decisions, with the invariance guarantees unchanged.
\end{abstract}

%% file: sections/model.tex
\paragraph{Typed questions.} A \emph{case} is one state $s$ (text or JSON) and $K$ questions. Each question has
instructions, a type $t\in\{\textit{choice},\textit{noul},\textit{score}\}$ and options $o_1,\dots,o_n$, each written with
its criterion description (``\texttt{human\_review}: queue this trace for a human'', ``level 3: critical; act the same
day'', ``true: a human should inspect this run''). The model returns a distribution over the declared options of every
question.

\paragraph{Packed layout.} Verdict-MQ reads a whole case in one pass of a bidirectional encoder:
{\small
\begin{align*}
&\texttt{[CLS]}\; s\; \texttt{[SEP]}\;\|\; b_1 \;\|\; \dots \;\|\; b_K, \\
&b_k = \langle t_k\rangle\, q_k\;\texttt{[SEP]}\;\texttt{[OPT]}\,o_{k,1}\,\dots\,\texttt{[OPT]}\,o_{k,n_k},
\end{align*}}
where $\langle t_k\rangle$ is the rendered prefix ``\textit{choice} question:'' (or \textit{noul}, \textit{score}) and $q_k$ the
question's instructions.

\input{figures/mask}

\paragraph{Structured attention.} One boolean mask (Figure~\ref{fig:mask}), applied identically in ModernBERT's global and
sliding-window layers, enforces four rules:
\begin{enumerate}
\item \emph{State tokens attend only to state tokens.} The state encoding is independent of the questions: no question
can change it, and it can be cached. This holds \emph{for identical retained state tokens}: packing gives the state
$\text{max\_len}$ minus the questions' length, so a long state can be truncated more when packed with more
questions. At $\text{max\_len}{=}2048$ no evaluated test case was truncated (worst-case margin 567 tokens;
Appendix~\ref{app:hparams}), so no reported number is affected.
\item \emph{Block $b_k$ attends to the state and to itself}, and every block starts at the position right after the
state. Answers do not depend on which other questions share the pass or in what order; packed answers equal
single-question answers (same precondition as above).
\item \emph{Inside a block, question tokens attend to every option, while each option's tokens attend only to the state,
the question and their own option.} Alternatives reach an option through the question tokens.
\item \emph{All options of a block share the same position ids}, and the sliding window is defined on position ids rather
than sequence index. Permuting options permutes the outputs and changes nothing else.
\end{enumerate}

\paragraph{Option marker.} Each option starts with a marker token, written \texttt{[OPT]} above, whose final hidden state
is read out. Reusing an existing special token as the marker, rather than a newly added token with a randomly
initialized embedding, raises the large model's typed-decisions accuracy from
0.790\,$\pm$\,0.007 to 0.801\,$\pm$\,0.005 over five fresh seeds each (Welch's $t$ over seeds, $p=0.022$;
Section~\ref{sec:typed}). All headline models
use \texttt{[CLS]} as the marker.

\paragraph{Readout and training.} The final hidden state at each option marker, plus a question-type embedding,
passes through a two-layer MLP to a logit; a softmax over the declared options gives the distribution. Training minimizes
cross-entropy against the benchmark's soft label distributions \citep{peterson2019human} plus a squared earth-mover term
\citep{hou2016emd} on \textit{score} questions; choice questions with more than two options are trained on random option
subsets (probability 0.3, gold kept). An optional \emph{option mixer} inserts a small Transformer encoder \emph{without
positional encodings} over the option markers of each question before the MLP. Self-attention over an unordered set is
permutation-equivariant \citep{zaheer2017deepsets,lee2019settransformer}, so options can be compared with each other while
rule 4's guarantee is kept.

\paragraph{Verified properties.} With a real ModernBERT backbone and a randomly initialized head, permuting options,
answering a question alone rather than packed, and batching with padding each change output probabilities by at most
$5.4\times10^{-7}$ in fp32 ($2.1\times10^{-6}$ with full-depth split); the guarantee is exact in exact arithmetic,
and bf16, used for all reported numbers, loosens the realized bound to $3.9\times10^{-3}$ (worst case over 20 random
cases). On a trained checkpoint, permutation never changes the chosen option
or crosses the autonomy threshold; answering alone never changes the chosen option but crosses the threshold in
0.75\% of cases, and bf16 vs fp32 flips 0.3\% of decisions (Appendix~\ref{app:hparams}). Option markers remain
distinct (minimum pairwise L2 distance 6.2), and fifteen gradient steps on one case raise every gold probability by at
least 0.5.

\paragraph{A failure mode.} If every option may attend to every other option, the markers are identical tokens at
identical positions with identical visibility, so they collapse to one representation: the model outputs uniform
distributions and does not learn, while invariance tests pass trivially. Rule 3 prevents this, and
our tests check marker distinctness and learnability as well as invariance.

\paragraph{Cost.} One pass over $|s|+\sum_k|b_k|$ tokens replaces $K$ passes over $|s|+|b_k|$ tokens.

\paragraph{Per-question baseline (Verdict-CE).} Verdict-CE renders one question with its options after the state and runs
one cross-encoder pass per question, scoring each option marker with the same MLP; it is the direct analogue of
per-question typed-decision models.

%% file: figures/mask.tex
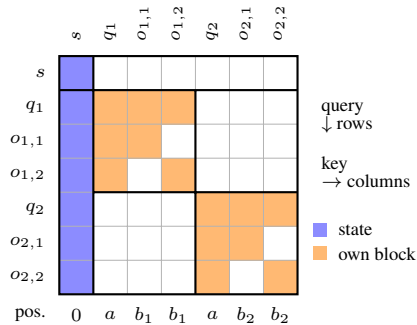
\begin{figure}[t]
\centering
\begin{tikzpicture}[x=0.45cm, y=0.45cm, every node/.style={font=\scriptsize}]
  \foreach \lab [count=\i from 0] in {$s$, $q_1$, $o_{1,1}$, $o_{1,2}$, $q_2$, $o_{2,1}$, $o_{2,2}$} {
    \node[anchor=east] at (-0.1, -\i-0.5) {\lab};
    \node[anchor=south, rotate=90, anchor=west] at (\i+0.5, 0.15) {\lab};
  }
  \foreach \r/\c in {0/0, 1/0, 2/0, 3/0, 4/0, 5/0, 6/0}
    \fill[blue!45] (\c, -\r) rectangle ++(1, -1);
  \foreach \r/\c in {1/1, 1/2, 1/3, 2/1, 2/2, 3/1, 3/3, 4/4, 4/5, 4/6, 5/4, 5/5, 6/4, 6/6}
    \fill[orange!55] (\c, -\r) rectangle ++(1, -1);
  \draw[step=1, gray!60, very thin] (0, 0) grid (7, -7);
  \draw[thick] (0, 0) rectangle (7, -7);
  \draw[thick] (1, 0) -- (1, -7) (0, -1) -- (7, -1) (4, -1) -- (4, -7) (1, -4) -- (7, -4);
  \node[anchor=east] at (-0.1, -7.6) {pos.};
  \foreach \p [count=\i from 0] in {$0$, $a$, $b_1$, $b_1$, $a$, $b_2$, $b_2$}
    \node at (\i+0.5, -7.6) {\p};
  \node[anchor=west, align=left, font=\scriptsize] at (7.4, -1.8) {query\\[-1pt]$\downarrow$ rows};
  \node[anchor=west, align=left, font=\scriptsize] at (7.4, -3.4) {key\\[-1pt]$\rightarrow$ columns};
  \fill[blue!45] (7.45, -4.9) rectangle ++(0.4, -0.4); \node[anchor=west] at (7.9, -5.1) {state};
  \fill[orange!55] (7.45, -5.6) rectangle ++(0.4, -0.4); \node[anchor=west] at (7.9, -5.8) {own block};
\end{tikzpicture}
\caption{Attention mask for a state $s$ and two questions with two options each (shaded: the row token attends to
the column token), with each group's position ids; it implements rules 1--4. Every block starts at position $a$ and a
block's options share one start position $b_k$, so neither question grouping nor option order is visible.}
\label{fig:mask}
\end{figure}

%% file: sections/setup.tex
\paragraph{Benchmark.} Typed-decisions \citep{typeddecisions2026} has four synthetic workflows (agent-trace observability,
customer service, invoice processing, security incidents), 1,200 training cases (6,000 decisions) and 400 test cases
(2,000 decisions), five typed questions per case. Gold labels are the mean of three samples from a teacher of roughly 4B-class
capability, so scores measure agreement with that teacher; the card reports a ``perfect scenario understanding'' ceiling
of 0.704 and teacher self-agreement of 0.735, and calls about 0.75 saturation: a score much above it ``means a model has
learned the teacher's quirks rather than the task.''
Every result in Sections~\ref{sec:typed}--\ref{sec:attribution} is in the card's \emph{specialist} mode (trained on its
four workflows) unless marked.

\paragraph{Metrics.} We report the card's metrics (defined in Table~\ref{tab:typed-main}), including 10-bin top-label
ECE \citep{naeini2015ece}. Our implementation reproduces the card's input-ignoring Uniform row exactly on KL (0.444), TV
(0.381) and Brier (0.238); the card does not specify its ECE binning or prior construction, so ECE and the prior are
compared only among systems we measured.

\paragraph{Baselines.} Laya's released checkpoints are measured by us with the benchmark's native question definitions,
criteria included \citep{laya2026}; per its model card, the typed-decisions checkpoint is fine-tuned on the benchmark's
training split, a like-for-like specialist. Jev is likewise measured by us, through its public API on Vercel's AI Gateway
(\texttt{typesafe-ai/jev}), one call per case with all five questions batched; it has no specialist mode available to
us, so it is scored zero-shot like the card. On RACE, BoolQ and MultiRC both serve as zero-shot references
(Section~\ref{sec:standard}). Decider, Featherless's Simple Jev and the card's
per-workflow ModernBERT-base and MiniLM classifiers are taken from the benchmark card; we did not re-run them. Verdict-CE is our per-question cross-encoder.

\paragraph{Protocol.} All hyper-parameter and recipe choices are made on a development split: 15\% of the training cases,
held out by a hash of the state. The last epoch is always reported; no checkpoint is chosen on the test split, though
the research process as a whole was adaptive rather than test-blind (Limitations). Headline
cells use three or more seeds, and every seed of a chosen recipe is reported. Comparisons with Laya use a 10,000-sample
case-cluster bootstrap \citep{efron1993bootstrap} as the primary significance test; item-level McNemar
\citep{mcnemar1947} is reported alongside but is anti-conservative, since one case's several decisions are correlated.

\paragraph{Training.} Full hyperparameters (optimizer, schedule, learning rates, epochs) are in
Appendix~\ref{app:hparams}, Table~\ref{tab:hparams}. Initializations: ModernBERT-base/large
\citep{warner2024modernbert}, the multi-task NLI versions released by tasksource
\citep{sileo2024tasksource}\footnote{\url{https://huggingface.co/tasksource/ModernBERT-base-nli} and
\url{https://huggingface.co/tasksource/ModernBERT-large-nli}.}, and Laya's own encoder. All runs use one RTX 4090
Laptop GPU (16~GB).

%% file: sections/typed.tex
\input{tables/typed_main}

Table~\ref{tab:typed-main} gives the main comparison; we read accuracy and the distribution metrics separately. \emph{Accuracy} is
agreement with a small teacher's argmax; the card calls about 0.75 saturation and reads scores much above it as learning
the teacher's quirks. Laya (0.767) is at that point and our large model (0.801) above it, so we treat accuracy margins
among them as closer agreement with the teacher, not as evidence of better decisions. The \emph{distribution} metrics against the soft gold, which the
card recommends over ECE (an input-ignoring prior has the best ECE on its table), are where the systems differ. The
150M NLI-initialized model ties Laya on accuracy (0.766\,$\pm$\,0.008 vs 0.767) and is better on every distribution
metric; its KL and TV gains are significant on all three seeds, but under soft-gold temperature matching they shrink (KL by
40--55\%, TV by about 70\%) and remain significant on two of three (Appendix~\ref{app:calib-methods}).

\paragraph{Large model and training stability.} The 395M model first trained unstably: with the development-selected
6\% warmup, two of six seeds stalled (0.707, 0.658), a known issue for large encoders
\citep{dodge2020finetuning,mosbach2021stability}. After comparing two standard remedies on the stalled seeds' training
loss and then on the development split, a pre-registered rule chose 20\% warmup, which reaches 0.790\,$\pm$\,0.007
over five fresh test seeds without stalls. Replacing the new marker token with an existing one
(Section~\ref{sec:model}) reaches our headline \textbf{0.801\,$\pm$\,0.005} over five more fresh seeds: closer
agreement with the teacher's labels than Laya on every seed (case-cluster bootstrap $p\le0.0018$, four of five at the
resolution floor) and a significantly better match to its soft labels (Table~\ref{tab:paired}). Reading each question
in the top 22 of the model's 28 layers ($k$=22) gives 0.804\,$\pm$\,0.003, statistically indistinguishable (Welch's
$t$ over seeds, $p=0.24$), so we keep the shared $k$=0 setting; Appendix~\ref{app:paired-full} lists every intermediate recipe and seed.

\input{tables/typed_paired}

\paragraph{Temperature-matched comparison.} One scalar temperature can move probabilistic metrics without
changing what a system knows, so we refit each system's temperature on held-out cases. Fitted to the hard labels, both
calibrate well (ECE: Laya 0.030, Verdict-MQ 0.017--0.032) and the residual ECE difference is not statistically
established. Fitted to the soft gold instead (Laya 0.78, Verdict-MQ about 1.13), Verdict-MQ's held-out KL stays well
below Laya's (0.074--0.077 vs 0.105), as does its Brier, significantly on every seed (Table~\ref{tab:paired}): it fits
the benchmark's label distributions better, not merely at a better temperature.

\paragraph{The autonomy dial.} An operator needs a confidence threshold, chosen on past data, that delivers its target
on new data. Table~\ref{tab:autonomy} splits test \emph{cases} in half by a hash shared by all systems, picks each
system's threshold on one half from raw confidence, and scores the other; the target is agreement with the gold labels,
i.e.\ with the teacher. At a 90\% target the large Verdict-MQ auto-handles 62--71\% of decisions against Laya's 56\%,
at comparable or better realized accuracy (89.4--92.2\% vs 86.5\%), with 32--39\% lower AURC. The gain is in how
decisions are \emph{ranked} by confidence and does not extend to ECE-style calibration: the headline model's raw ECE (0.156) is worse
than the 150M model's, which covers fewer decisions than Laya. On temperature-calibrated confidence Verdict-MQ stays
ahead on AURC, but strict dominance at 90\% fails for 2 of 5 seeds. Coverage is per decision: auto-handling a whole
case only once all five decisions clear $\tau$ qualifies just 19--27\% of cases (Laya 2.5\%;
Appendix~\ref{app:calib-methods}).

\input{tables/autonomy}

\paragraph{Latency.} Table~\ref{tab:latency} measures the same weights in packed and one-question-per-pass mode. Packing
the five questions of a case is 2.7$\times$ faster per case for the base model and 2.8$\times$ for the large model at batch
size 8, and 1.3--1.8$\times$ at batch size 1, where the GPU is under-utilized in both modes.

\input{tables/latency}

%% file: tables/typed_main.tex
\begin{table*}[t]
\centering
\small
\begin{adjustbox}{max width=\textwidth}
\begin{tabular}{lrrrrrrrrr}
\toprule
System & Acc & Soft & F1 & KL & TV & Brier & ECE & MAE & W1 \\
\midrule
Uniform (card) & 0.308 & 0.311 & 0.152 & 0.444 & 0.381 & 0.238 & 0.169 & -- & -- \\
Prior (card) & 0.470 & 0.430 & 0.207 & 0.347 & 0.317 & 0.189 & 0.088 & -- & -- \\
MiniLM-L6 per workflow (card) & 0.587 & 0.506 & 0.424 & 0.262 & 0.267 & 0.143 & 0.108 & 0.515 & 0.864 \\
ModernBERT-base per workflow (card) & 0.646 & 0.542 & 0.469 & 0.223 & 0.249 & 0.119 & 0.179 & 0.444 & 0.931 \\
Featherless Simple Jev (card) & 0.716 & -- & -- & 0.488 & -- & 0.176 & -- & -- & -- \\
TypeSafe Jev 1.13.0 (card) & 0.727 & 0.580 & 0.613 & 1.442 & 0.251 & 0.148 & 0.144 & 0.391 & 0.952 \\
meraGPT Decider 1 (card) & 0.768 & 0.608 & 0.641 & 0.096 & 0.149 & 0.052 & 0.180 & 0.219 & 0.984 \\
\midrule
Laya general checkpoint, zero-shot (measured) & 0.358 & 0.347 & 0.239 & 0.570 & 0.410 & 0.316 & 0.178 & 0.695 & 0.726 \\
Laya typed-decisions checkpoint, 421M (measured) & 0.767 & 0.605 & 0.661 & 0.117 & 0.174 & 0.061 & 0.214 & 0.242 & 0.981 \\
TypeSafe Jev, general (measured, Appendix~\ref{app:reproduce-jev}) & 0.734 & 0.583 & 0.639 & 0.829 & 0.251 & 0.148 & 0.028 & 0.391 & 0.966 \\
\midrule
Verdict-CE, per-question cross-encoder, 150M & 0.586 & 0.499 & 0.440 & 0.245 & 0.255 & 0.141 & 0.066 & 0.455 & 0.875 \\
Verdict-MQ, ModernBERT-base init, 3 seeds & 0.681\,$\pm$\,0.013 & 0.553 & 0.545 & 0.191\,$\pm$\,0.009 & 0.213 & 0.108\,$\pm$\,0.006 & 0.095\,$\pm$\,0.010 & 0.364 & 0.934 \\
Verdict-MQ, NLI-base init, 150M, 3 seeds & 0.766\,$\pm$\,0.008 & 0.602 & 0.660 & 0.103\,$\pm$\,0.001 & 0.149 & 0.057\,$\pm$\,0.001 & 0.135\,$\pm$\,0.008 & 0.241 & 0.977 \\
Verdict-MQ, NLI-large init, 395M, 5 fresh seeds & 0.801\,$\pm$\,0.005 & 0.616 & 0.702 & 0.080\,$\pm$\,0.001 & 0.129 & 0.045\,$\pm$\,0.001 & 0.156\,$\pm$\,0.007 & 0.200 & 0.983 \\
\bottomrule
\end{tabular}
\end{adjustbox}
\caption{Typed-decisions test split (2,000 decisions, 400 cases), specialist mode unless marked. Acc = accuracy; Soft = soft accuracy; F1 = macro-F1; KL, TV, Brier against the benchmark's soft gold (lower is better); ECE 10-bin; MAE = score-question MAE of the expected level; W1 = score predictions within one level; $\pm$ = seed standard deviation. Card rows are taken from the benchmark card (its Jev row is the card maintainer's own measurement); we additionally measure Jev ourselves (387/400 cases; 13 excluded for transient errors), closely matching it. Jev, Simple Jev and Decider are general (zero-shot), not directly comparable with specialists; Simple Jev and Decider are card-only (the card gives only Acc, KL and Brier for Simple Jev). The card calls accuracy of about 0.75 saturation. Intermediate training-recipe variants (marker and split-depth ablations) are Table~\ref{tab:typed-main-full} in the appendix.}
\label{tab:typed-main}
\end{table*}

%% file: tables/typed_paired.tex
\begin{table*}[t]
\centering
\small
\begin{adjustbox}{max width=\textwidth}
\begin{tabular}{lrrrrrr}
\toprule
Seed & Acc & $\Delta$Acc vs Laya [95\% CI] & $p$ & $\Delta$KL [95\% CI] & $\Delta$Brier [95\% CI] & $\Delta$KL, temp.-matched [95\% CI] \\
\midrule
Seed 40 & 79.8 & $+3.1$ [$+1.4$, $+4.8$] & 0.0002 & $-0.036$ [$-0.042$, $-0.031$] & $-0.017$ [$-0.020$, $-0.014$] & $-0.028$ [$-0.035$, $-0.022$] \\
Seed 41 & 80.3 & $+3.6$ [$+1.9$, $+5.4$] & 0.0002 & $-0.036$ [$-0.042$, $-0.030$] & $-0.017$ [$-0.020$, $-0.013$] & $-0.030$ [$-0.037$, $-0.023$] \\
Seed 42 & 80.8 & $+4.0$ [$+2.4$, $+5.7$] & 0.0002 & $-0.038$ [$-0.044$, $-0.032$] & $-0.018$ [$-0.022$, $-0.015$] & $-0.029$ [$-0.036$, $-0.023$] \\
Seed 43 & 80.0 & $+3.2$ [$+1.6$, $+4.9$] & 0.0002 & $-0.038$ [$-0.043$, $-0.032$] & $-0.017$ [$-0.021$, $-0.014$] & $-0.031$ [$-0.037$, $-0.024$] \\
Seed 44 & 79.5 & $+2.8$ [$+1.2$, $+4.5$] & 0.0018 & $-0.035$ [$-0.041$, $-0.029$] & $-0.015$ [$-0.019$, $-0.012$] & $-0.029$ [$-0.036$, $-0.022$] \\
\bottomrule
\end{tabular}
\end{adjustbox}
\caption{The five fresh headline seeds (395M, fixed marker) against the released Laya typed-decisions checkpoint on the identical 2,000 test decisions, all paired by case: case-cluster bootstrap over the 400 cases (10,000 resamples). $\Delta$ = Verdict-MQ $-$ Laya, so negative KL and Brier (against the soft gold) favour Verdict-MQ. Temp.-matched: one temperature per system, fitted to the soft gold on half of the cases, scored on the other half. Every KL and Brier difference here, raw or temperature-matched, has $p\le0.0002$. The full search history, with item-level McNemar and $\Delta$ECE, is Table~\ref{tab:paired-full}.}
\label{tab:paired}
\end{table*}

%% file: tables/autonomy.tex
\begin{table*}[t]
\centering
\small
\begin{adjustbox}{max width=\textwidth}
\begin{tabular}{lrrrrr}
\toprule
System & Acc & AURC & 90\% target & 95\% target & 98\% target \\
\midrule
Laya & 76.7\% & 0.1069 & 55.7\% / 86.5\% & 36.8\% / 94.5\% & 20.9\% / 99.5\% \\
Verdict-MQ large s40 & 79.8\% & 0.0698 & 67.3\% / 89.4\% & 43.7\% / 96.3\% & 24.3\% / 99.2\% \\
Verdict-MQ large s41 & 80.3\% & 0.0665 & 65.0\% / 91.7\% & 42.4\% / 96.2\% & 29.0\% / 99.0\% \\
Verdict-MQ large s42 & 80.8\% & 0.0655 & 71.1\% / 90.4\% & 51.2\% / 94.9\% & 26.4\% / 99.2\% \\
Verdict-MQ large s43 & 80.0\% & 0.0670 & 64.0\% / 92.2\% & 41.2\% / 96.8\% & 25.1\% / 99.2\% \\
Verdict-MQ large s44 & 79.5\% & 0.0725 & 62.2\% / 91.8\% & 43.7\% / 95.6\% & 25.6\% / 99.6\% \\
Verdict-MQ base s0 & 76.3\% & 0.1024 & 40.3\% / 93.8\% & 25.7\% / 97.7\% & 19.7\% / 99.0\% \\
Verdict-MQ base s1 & 77.5\% & 0.0900 & 42.8\% / 94.8\% & 28.0\% / 98.6\% & 20.2\% / 99.0\% \\
Verdict-MQ base s2 & 76.1\% & 0.0968 & 52.9\% / 90.7\% & 27.1\% / 98.1\% & 22.5\% / 99.6\% \\
\bottomrule
\end{tabular}
\end{adjustbox}
\caption{Held-out autonomy dial on typed-decisions test decisions. The confidence threshold for each target accuracy is chosen on one half of the test \emph{cases} (split by a fixed hash shared by all systems, so one case's 5 decisions never straddle the split) and applied to the other half; each cell gives coverage (share auto-handled) / realized accuracy on that half. AURC = area under the risk-coverage curve (lower is better).}
\label{tab:autonomy}
\end{table*}

%% file: tables/latency.tex
\begin{table}[t]
\centering
\small
\begin{adjustbox}{max width=\columnwidth}
\begin{tabular}{lrrrr}
\toprule
Model & Batch & Packed & Per question & Speed-up \\
\midrule
Verdict-MQ base & 1 & 9.6 & 12.5 & 1.3$\times$ \\
Verdict-MQ base & 8 & 5.8 & 15.4 & 2.7$\times$ \\
Verdict-MQ large & 1 & 13.4 & 24.5 & 1.8$\times$ \\
Verdict-MQ large & 8 & 11.4 & 32.0 & 2.8$\times$ \\
\bottomrule
\end{tabular}
\end{adjustbox}
\caption{Latency in ms per case: all five questions in one packed pass vs one pass per question (typed-decisions test cases; bf16; RTX 4090 Laptop GPU; same weights in both modes; measured on checkpoints with the original marker token, whose compute is identical to the headline models').}
\label{tab:latency}
\end{table}

%% file: sections/attribution.tex
The previous section compares complete systems. Here we ask what the architecture itself contributes, separately from the
initialization and the training recipe. Our principal architectural evidence is the fully controlled $k$=0 vs.\ $k$=22
comparison in Section~\ref{sec:standard} (same initialization, loss, training objective, epochs and learning rate,
chosen from the same pre-registered search, differing only in whether the state can read its question): it shows the
architecture's shared-state design costs 0.9--3.5 points of accuracy on RACE and MultiRC relative to letting the state
attend to its question. The invariance guarantees are identical in both arms, since at $k>0$
each question gets its own copy of the state; the comparison measures the cost of sharing, not of invariance. Three further checks are weaker evidence, since none isolates the
architecture as cleanly: (i) at a matched starting encoder but \emph{unmatched}
training recipes, the two architectures tie within about a point, so we do not claim the architecture is more accurate
than Laya's; (ii) component ablations show most of Verdict-MQ's add-ons (question-type embedding, shared positions,
ordinal loss, long context, the option mixer) do not improve accuracy on this benchmark and are kept for the invariance
guarantee or reported as negative results; (iii) against the per-question cross-encoder, only the permutation-robustness difference is
architectural by construction (rules 3 and 4); the overall accuracy gap is weaker evidence, since training details
differ. Full numbers are in Appendix~\ref{app:attribution-methods}, including a learning-rate ablation that looked
better on the test split, was re-checked on development, and was not adopted.

The same question applies to Section~\ref{sec:typed}'s distributional (KL, Brier) advantage: training on the soft gold
could win it from the objective alone, but raw-ModernBERT-initialized Verdict-MQ, trained the same way, does not show
it (KL 0.191 at 150M, 0.153 at 395M, both worse than Laya's 0.117; Table~\ref{tab:typed-main-full}), so we read it too
as most consistent with the NLI initialization, not the architecture.

%% file: sections/standard.tex
Typed-decisions is small, synthetic and not peer-reviewed, so we also evaluate on three standard, peer-reviewed
benchmarks: RACE, BoolQ and MultiRC. Verdict-MQ and a baseline classifier with the same initialization and epoch
budget are both fine-tuned on each benchmark's training split. The two systems from Section~\ref{sec:related}, Laya's
general checkpoint and Jev, serve as \emph{zero-shot references}: we do not fine-tune them, and they are run on the same
test items and case grouping (Table~\ref{tab:standard}).\footnote{Jev drops 2.1--3.4\% of items to transient API errors;
a common-subset recheck confirms this favours no system (Appendix~\ref{app:standard-methods}).} Because the NLI
initialization was trained on SuperGLUE BoolQ and MultiRC, those two benchmarks use raw ModernBERT as the primary
initialization.

\input{tables/standard}

\emph{Configurations.} RACE and MultiRC use split depth $k$=22: in the top $k$ layers each question reads its own copy
of the state, and at $k$=22, the full depth of ModernBERT-base, nothing is shared across questions. The invariance
guarantees hold at every $k$ (verified numerically); what shrinks as $k$ grows is the sharing, and with it the speed-up
and the cacheable state. BoolQ uses $k$=0.

\emph{Matched comparison.} Against the matched fine-tuned baseline, Verdict-MQ wins on MultiRC ($+9$ points over the
baseline's mean, F1a 77.5 vs 67.8; the baseline varies widely across seeds, and the margin is 5--8 points against its
two converged seeds and 3 points against the best run of its learning-rate sweep), trails narrowly on RACE ($-1.5$
points, consistent across seeds), and trails on BoolQ on all five seeds ($-1.6$ to $-9.3$ points, $p\le0.025$). The
zero-shot systems are reference points rather than competitors. Laya's general checkpoint is far below every
fine-tuned model on RACE and MultiRC but matches Verdict-MQ on BoolQ (76.0 vs 76.2\%), so on BoolQ fine-tuning
Verdict-MQ yields no gain over it; a fine-tuned Laya, the like-for-like comparison, was not run. Jev, very likely built
on a much larger base model, outperforms every fine-tuned encoder (90--93\%).

\emph{Efficiency.} Every local encoder (Verdict-MQ at 8.5--51~ms per case, the baseline at 7--50~ms;
Table~\ref{tab:split}) is one to two orders of magnitude faster than Jev's API, about 35--205$\times$ excluding our
own 0.3~s inter-call delay, or 14--84$\times$ against the card's own 0.71~s median (Appendix~\ref{app:standard-methods}). This
reflects local inference, not our architecture: against the matched baseline, Verdict-MQ is faster only on RACE (17.2
vs 49.6~ms) and slower on BoolQ (8.5 vs 7.2) and MultiRC (50.9 vs 44.0). Its own efficiency shows in packing
(Table~\ref{tab:latency}) and at $k$=0, where it beats the baseline on MultiRC in both latency (20.6 vs 44.0~ms) and
accuracy.

\emph{Controlled $k$=0 vs.\ $k$=22 comparison.} $k$=0 was one of four complete candidates in the pre-registered
development-split search that selected the $k$=22 recipe; the two share initialization, loss, objective, epochs and
learning rate, and differ only in $k$ (Appendix~\ref{app:standard-methods}). One of the three $k$=0 test seeds reuses
the search seed's previously reported accuracy rather than being fresh (Appendix~\ref{app:standard-methods}). Letting
the state read its question yields a modest accuracy gain: on RACE, $k$=22 reaches 77.4\,$\pm$\,0.4\% against
76.2\,$\pm$\,0.2\% at $k$=0 ($+1.0$ to $+1.5$ points per paired seed, $p$ 0.006--0.045); on MultiRC,
80.6\,$\pm$\,0.5\% against 78.4\,$\pm$\,0.8\% ($+0.9$ to $+3.5$, $p$ 0.0002--0.11, significant on two of three seeds).
The gain costs 1.4$\times$ the per-case latency on RACE (17.2 vs 12.1~ms) and 2.5$\times$ on MultiRC (50.9 vs
20.6~ms) at batch size 1 (seed 0, Table~\ref{tab:split}), while the invariance guarantees are identical in both
configurations. At $k$=0, Verdict-MQ still beats the matched baseline on MultiRC on every seed ($+1.7$ to $+12.8$
points, all significant). On RACE the gap to the baseline persists at full depth ($-1.5$ points at $k$=22), so the
shared state does not explain it; its cause remains open.

\emph{BoolQ stability.} BoolQ training is unstable across seeds. With the marker fix (Section~\ref{sec:model}), roughly
half of runs converged and half stalled near the majority-class rate (62\%). We therefore selected the recipe by
the stability of development accuracy across seeds, as an early-loss criterion proved unreliable; the chosen recipe's
development seeds lie within 1.4 points of each other, yet five fresh test seeds range from 72.3 to 79.4\%. Verdict-MQ
trails the epoch-matched baseline on BoolQ, and a single training run's accuracy there is uncertain by several points.

%% file: tables/standard.tex
\begin{table*}[t]
\centering
\small
\begin{adjustbox}{max width=\textwidth}
\begin{tabular}{lrrrrr}
\toprule
Benchmark & Verdict-MQ & Baseline & $p$ vs baseline & Laya, zero-shot & Jev, zero-shot \\
\midrule
RACE (test) & 77.4$\pm$0.4 & 78.9$\pm$0.1 & 0.0026--0.021 & 45.5 & 93.3 \\
BoolQ (val)$\dagger$ & 76.2$\pm$3.1 & 81.0$\pm$0.6 & 0.0002--0.025 & 76.0 & 91.3 \\
MultiRC (val)$\ddagger$ & 80.6$\pm$0.5 & 71.6$\pm$4.9 & 0.0002--0.0002 & 54.9 & 90.0 \\
\bottomrule
\end{tabular}
\end{adjustbox}
\caption{Standard benchmarks. Verdict-MQ and the baseline (same initialization and epoch budget) are fine-tuned on each benchmark's training split; Laya's general checkpoint and Jev are zero-shot references (not fine-tuned on these tasks; their own training data is undisclosed), run on the identical test items and case grouping. All numbers are measured by us, not taken from vendor cards (Appendix~\ref{app:standard-methods}). $p$: case-cluster bootstrap, min--max across seeds (3 for RACE/MultiRC, 5 for BoolQ). $\dagger$BoolQ training is unstable (\S\ref{sec:standard}): the 5 seeds range 72.3--79.4\%. $\ddagger$MultiRC official metrics: Verdict-MQ F1a/EM 77.5/38.8 vs baseline 67.8/20.7; the baseline varies widely across seeds (66.4, 75.9, 72.6\%; Appendix~\ref{app:standard-methods}).}
\label{tab:standard}
\end{table*}

%% file: sections/conclusion.tex
Verdict-MQ reads a case once and answers all its typed questions, invariant by construction to option order and
question grouping given identical retained state tokens (Section~\ref{sec:model}). On typed decisions it matches the
soft labels significantly better than Laya's specialist on every seed, robust to temperature scaling, while accuracy
there is at or past saturation (Section~\ref{sec:typed}). On standard benchmarks it beats a matched
fine-tuned baseline on MultiRC and loses on RACE and BoolQ, and a zero-shot closed system beats every fine-tuned
encoder (Section~\ref{sec:standard}, one to two orders of magnitude slower). The controlled comparison
quantifies the shared state's cost: 0.9--3.5 points on RACE/MultiRC for 1.4--2.5$\times$ the latency, nothing on
typed decisions. Code and seeds are released at \coderepo.

%% file: sections/limitations.tex
\paragraph{Benchmarks.} Typed-decisions is small (400 test cases), synthetic, labelled by a small teacher model and not
peer-reviewed, so high scores partly measure agreement with that teacher. The standard benchmarks
(Section~\ref{sec:standard}) complement it, but our headline typed-decision results for Verdict-MQ are specialist results; the generalists we trained
(Appendix~\ref{app:generalist}) fall below the input-ignoring prior.

\paragraph{Reading comprehension and stability.} On RACE, Verdict-MQ remains 1.5 points below the matched baseline
even at $k$=22, where each question reads its own copy of the state, so this gap is not explained by the shared state
and its cause is open. BoolQ required several successive recipe changes before training stopped failing on about half of
runs, and the final recipe still spans 7 points across test seeds; runs at $k>0$ were also unstable across seeds
(Appendix~\ref{app:split}), so the instability is not specific to the shared state. Until it is understood, Verdict-MQ
is not a reliable choice for BoolQ-like tasks.

\paragraph{Comparisons.} The comparison with an open system is like-for-like only on typed decisions, against Laya's
released specialist, to our knowledge the only open typed-decision model of this size. On RACE, BoolQ and MultiRC,
Laya's general checkpoint and Jev are zero-shot references; we did not fine-tune Laya there and make no claim about a
fine-tuned Laya on those tasks. Jev, a general and likely much larger closed system, outperforms every model here on
those benchmarks. The MultiRC baseline varies widely across seeds, and we did not run a stability search for it
comparable to the one for Verdict-MQ on BoolQ.

\paragraph{Selection and testing.} All final configuration choices (option marker, split depth, BoolQ recipe) were made
on training loss or a held-out development split, never on the reported test split. The research process was
nonetheless adaptive rather than test-blind, in three places. (i) A constant-learning-rate ablation looked better on
the raw-ModernBERT test split (0.733 vs 0.681); we re-checked it on development, where it did not hold, and did not
adopt it, but the re-check was prompted by the test result (Appendix~\ref{sec:schedule}). (ii) The five-seed
large-model run at the selected (marker, $k$=22) configuration was launched only if the base model's test accuracy
there had not regressed by more than 0.5 points against an earlier base result; the $k$=0 headline seeds were run
afterwards, with those test results in hand. (iii) Every typed-decisions training run also prints test accuracy after
each epoch, mislabelled ``val'' (the scripted recipe searches in Appendix~\ref{app:attribution-methods} use a
genuinely held-out slice of the training cases instead, and never read this print, but a
human monitoring training saw live test accuracy on every run, not just (i)-(ii); these per-run logs are not
released). Test examples never entered training, and reported numbers come from
seeds not previously evaluated on test at their configuration, with one disclosed exception: the $k$=0
standard-benchmark comparison reuses one previously reported seed (Appendix~\ref{app:standard-methods}). Reported
$p$-values are not adjusted for the seeds, benchmarks or configurations tried, and the test behind each is named
where it is reported (case-cluster bootstrap, McNemar, or Welch's $t$ over seeds), so a $p$-value indicates how
surprising a final result is, not how many alternatives preceded it.

\paragraph{Uncontrolled differences.} The large model trains stably only with a 20\% warmup, and because our scheduler
applies the warmup on top of the decay, a longer warmup also lowers the realized peak learning rate (from about 94\% of
the configured rate at 6\% warmup to about 80\% at 20\%); without a control that holds the peak fixed, we cannot
attribute the stability to warmup length rather than to the lower peak. Packing also changes the loss weighting
relative to the baseline: our loss averages over the questions of a case and then over cases, so a question in a
large case (MultiRC averages 14.1 questions per packed case, up to 16; RACE 3.5; BoolQ 1.1) receives less gradient than
a question alone in its case, whereas the baseline weights all decisions equally. This matters most for MultiRC, and we
did not test a per-decision-weighted loss. At matched initialization Verdict-MQ ties Laya's architecture on accuracy,
so our accuracy results describe a complete open system rather than the architecture alone; our re-fine-tuning of
Laya used hard labels without temperature scaling, which favours Verdict-MQ on distribution metrics in that comparison
only.

\paragraph{Scope of measurements.} Jev is called by the unpinned alias \texttt{typesafe-ai/jev}, so its numbers may not
be reproducible after a vendor update (Appendix~\ref{app:reproduce-jev} records the date and response metadata). Email
labels come from LLM teachers, not humans, and one planned email task proved unusable. All local timings come from one
laptop GPU, and Jev's per-call times include gateway overhead and retry backoff that we could not separate. We do not
yet report validation on a live organizational corpus. Autonomy coverage is per decision: under a strict policy that
automates a case only when all its decisions clear the threshold, whole-case automation is roughly a third of the
per-decision figure, and the accuracy of fully automated cases can fall below the target.

%% file: sections/ethics.tex
\paragraph{Data.} All benchmarks are public research datasets (licences in Appendix~\ref{app:licences}). The
organizational-email experiments use the public Enron corpus \citep{klimt2004enron}; we release labels keyed to
message hashes, not message bodies. Email labels come from LLM teachers with no human annotators; their agreement and
limitations are reported in Appendix~\ref{app:email-full}.

\paragraph{Intended use.} Verdict-MQ is meant to route organizational decisions either to automation or to a human. Its
accuracy targets are agreement with teacher-derived labels, and per-decision coverage overstates whole-case automation
(Section~\ref{sec:typed}), so a deployment should validate its thresholds on its own data and keep human review for
consequential decisions.

\paragraph{AI assistance.} Significant portions of this project's experiment execution, debugging, statistical analysis
and drafting were carried out with the assistance of an AI coding agent (Claude, Anthropic), under human direction and
review at every decision point reported in Sections~\ref{sec:typed} and~\ref{sec:standard}; final experimental design
choices, the decision to report negative results, and all claims in this paper remain the authors' responsibility.

%% file: sections/appendix.tex
\section{Controlled-attribution methodology detail}
\label{app:attribution-methods}
Full numbers for the three weaker attribution checks summarized in Section~\ref{sec:attribution}.

\paragraph{Matched initialization.} Table~\ref{tab:matched} starts both architectures from Laya's own general-purpose
encoder and trains them on the same 6,000 decisions, but not with matched training recipes: Laya's architecture was
re-fine-tuned with our \texttt{finetune\_laya.py} (3 epochs, batch 4, lr $10^{-5}$), while Verdict-MQ used
\texttt{train\_mq.py}'s own recipe (6 epochs, batch-cases 4, lr $2\times10^{-5}$). Over two seeds each, Laya's
architecture reaches 0.774 and 0.767 (mean 0.771) and Verdict-MQ 0.767 and 0.758 (mean 0.763): a tie within about a
point, with Laya's architecture slightly ahead, but because the recipes differ we cannot rule out that the recipe,
not just the architecture, contributed to this small gap.

\input{tables/matched}

\paragraph{Option interaction (the option mixer).} Rule 3 isolates each option from its alternatives. The option
mixer restores comparison between options with two permutation-equivariant set-attention layers over the option
markers, keeping the order-invariance guarantee. On the development split it added one point to the NLI-base model
(0.773\,$\pm$\,0.008 vs 0.763\,$\pm$\,0.001) and was chosen by our selection rule, but the gain did not reach the
test split: 0.764\,$\pm$\,0.005 over three fresh seeds against 0.766\,$\pm$\,0.008 without it, every seed tied with
Laya. For the large model it was neutral on development (0.797 vs 0.798) and was not carried to test, and at the
matched Laya initialization one of its two seeds stalled (Table~\ref{tab:matched}). We report it as a tested idea
that does not help here. Because the standard-benchmark runs were configured from the development split before this
test result was known, they use the two-layer mixer; on typed-decisions it is neutral.

\paragraph{Component ablations.} Table~\ref{tab:ablations} removes one component at a time. The question-type
embedding, shared option positions, the ordinal loss and the long context window do not move accuracy beyond seed
noise on this benchmark; option-subset augmentation may add 1.5 points (about one seed standard deviation). Shared
positions are kept for the invariance guarantee, not for accuracy.

\input{tables/ablations}

\paragraph{The learning-rate schedule.}
\label{sec:schedule}
The one ablation that moved accuracy was the schedule: a constant $2\times10^{-5}$ learning rate gave 0.733 against
0.681 for warmup and decay, with raw ModernBERT. Because this was a test-split observation, we re-chose the schedule
on the development split before adopting it. With the NLI initializations it did not help: for the base model the
constant $10^{-5}$ schedule tied warmup and decay on development (0.764 vs 0.763) and then scored 0.720, 0.737 and
0.713 on three fresh test seeds, well below the warmup-and-decay recipe; for the large model warmup and decay
remained best on development (Table~\ref{tab:devgrid}). The headline recipes are therefore unchanged. We include
this episode because a test-split ablation read at face value would have changed the recipe for the worse.

\section{Calibration and autonomy methodology detail}
\label{app:calib-methods}
This section gives the full numbers and procedure behind Section~\ref{sec:typed}'s calibration and autonomy-dial
claims.

\paragraph{Matched temperature scaling (top-label calibration).} We fit a matched, per-system temperature by grid-search
NLL minimization against the hard labels on one half of the test cases (\texttt{scripts/\allowbreak calibration\_temp\_scale.py}, case-level split, same convention as the
autonomy dial) and evaluated ECE on the other half. Verdict-MQ's calibrated ECE (0.017--0.032 across five seeds) is
descriptively lower than Laya's (0.030) on 4 of 5 seeds (by 0.007--0.013); one seed (s44, 0.032) is not. A paired,
case-cluster bootstrap 95\% CI on the calibrated-ECE difference (matched by case, 199 held-out clusters per seed)
includes zero for all five seeds.

\paragraph{Distribution metrics under matched temperature.} \texttt{scripts/\allowbreak distribution\_significance.py} runs the
same paired case-cluster bootstrap (400 cases, 10,000 resamples, add-one corrected) on KL, Brier and TV against the soft
gold. Raw, all five headline seeds beat Laya on all three metrics at the bootstrap's resolution floor (Table~\ref{tab:paired};
TV $-0.043$ to $-0.046$). To rule out a temperature artefact, each system's temperature is instead fitted to the
\emph{soft} gold on the fit half (Laya 0.78; Verdict-MQ 1.12--1.14) and the held-out half is scored: Verdict-MQ's KL is
0.074--0.077 against Laya's 0.105, Brier 0.041--0.042 against 0.056, TV 0.128--0.131 against 0.158, every difference
at the floor. The hard-label temperature above must not be reused for these metrics: it sharpens both systems'
predictions and raises both systems' held-out KL against the soft gold to 0.31--0.34, several times their raw values. For the 150M NLI-base model the raw KL and TV gains
are significant on all three seeds (Brier on one); after the soft-gold temperature fit they shrink (KL by 40--55\%, TV by about
70\%), and KL and TV remain significant on two of three seeds, Brier on none.

\input{figures/rc}

\paragraph{Autonomy dial, full numbers.} At a 98\% target Verdict-MQ's thresholds deliver 99.0--99.6\% realized
accuracy while Laya's delivers 99.5\%. Its AURC (held-out half only) is 0.066--0.073 vs Laya's 0.107 raw
(32--39\% lower), and 0.067--0.072 vs Laya's 0.102 calibrated (29--34\% lower). Recomputed on temperature-calibrated
confidence, the strict coverage-\emph{and}-accuracy dominance over Laya at the 90\% target does not survive for 2 of 5
seeds (s42, s43): their calibrated realized accuracy (88.8\%, 88.9\%) falls fractionally below Laya's calibrated
89.0\% while still covering more decisions (71.1\%, 72.7\% vs 58.4\%), so for those two seeds the extra coverage is
paid for in accuracy. The base model's own AURC (0.090--0.102 held-out) is on par with Laya's 0.107.

\paragraph{Case-level (joint-acceptance) automation, full numbers.} Requiring all five of a case's decisions to
individually clear $\tau$ before auto-handling the whole case (\texttt{scripts/case\_level\_autonomy.py}), at the 90\%
per-decision target: only 19--27\% of Verdict-MQ's held-out cases qualify (Laya: 2.5\%, 5 of 199 cases), and the
realized accuracy of those fully-auto-handled cases (69.8--81.6\% across Verdict-MQ's seeds) is itself below the 90\%
target the per-decision threshold was fit to: five independent decisions each individually clearing a 90\%-accuracy
bar do not compose to 90\% whole-case accuracy, since case-level correctness requires every decision correct
simultaneously.

\section{Standard-benchmark methodology detail}
\label{app:standard-methods}
This section gives the full accounting behind several claims stated briefly in Section~\ref{sec:standard}.

\paragraph{BoolQ's baseline epoch budget.} Verdict-MQ's final BoolQ recipe trains for 6 epochs, and so does the
baseline in Table~\ref{tab:standard} (5 seeds, 81.0\,$\pm$\,0.6\%). The baseline is insensitive to this choice: at 3
epochs it reaches 81.1\,$\pm$\,0.9\%.

\paragraph{Jev's item exclusions and the common-subset check.} Jev drops 2.1--3.4\% of items to transient API errors
(RACE 168/4{,}934, BoolQ 70/3{,}270, MultiRC 159/4{,}848) and is scored on the remainder, while every local system
(Verdict-MQ, the baseline, Laya) scores 100\% of items, so ``identical test items'' is not quite exact for Jev. We
recomputed every system's accuracy restricted to exactly the items Jev succeeded on
(\texttt{scripts/common\_subset\_compare.py}): every system's common-subset accuracy is within 0.3 points of its
full-split number on all three benchmarks, so the exclusion (as expected of transient, non-adversarial failures)
does not favor any system and does not change any reported ranking or gap.

\paragraph{Bootstrap resolution floor.} Our case-cluster bootstrap uses 10,000 resamples per comparison. A raw
0-crossings count would imply an unsupported exact-zero $p$-value, so each one-sided tail fraction is add-one
continuity corrected ($1/10{,}001$ at the floor; \citealp{davison1997bootstrap}), and the doubled two-sided statistic
therefore cannot go below $2/10{,}001\approx0.0002$. This is a Monte Carlo estimate, not an exact randomization test,
and the correction avoids an unsupported exact-zero claim without certifying precision tighter than that floor.

\paragraph{Single-utterance intent classification (CLINC150, Banking77).} These are a different setting from the rest
of this paper (one question per input, with nothing packed over shared state), included only as a
sanity check that reading label names generalizes. Verdict-MQ ties a standard classifier head on both (CLINC150
90.1\% vs 89.4\%, $p=0.07$; Banking77 92.0\% vs 91.8\%, $p=0.77$) while requiring no new output layer for new
intents; its ECE is worse (0.082 vs 0.015 on CLINC150, 0.054 vs 0.017 on Banking77), consistent with the classifier
head's simpler decision surface calibrating more easily than reading a description of each label.

\paragraph{Jev latency accounting.} Our standard-benchmark Jev runs were sequential, and the timed window included a
fixed 0.3\,s delay we inserted after every call and the retry backoff on transient errors: total
wall-clock time per call was 2.0--2.1\,s (RACE 2839\,s/1{,}407 cases, BoolQ 5999\,s/2{,}938, MultiRC 725\,s/344).
Removing our own delay gives 1.7--1.8\,s per call, which still includes retry backoff and the shared, rate-limited
gateway's overhead; the per-call time did not scale with the benchmarks' very different passage lengths, consistent
with that overhead dominating. Against Verdict-MQ's reported settings (8.5, 17.2 and 50.9\,ms per case on BoolQ, RACE
and MultiRC) this is about 35--205$\times$. The benchmark card independently reports a direct-API median of 0.71\,s
per case on its typed-decisions inputs, which would give about 14--84$\times$. We did not isolate Jev's
inference time, so we claim only that local inference is one to two orders of magnitude faster per case, a gap every
local encoder here shares; \texttt{scripts/eval\_jev\_std.py} now records per-call latency excluding its own delay.

\paragraph{MultiRC baseline variance.} Across the whole learning-rate sweep, MultiRC pair-baseline seeds land either at
65.2--66.4\% accuracy (F1a 0.61--0.62) or at 72.6--77.8\% (F1a 0.70--0.74). At the dev-chosen learning rate
($10^{-5}$) the three seeds score 66.4, 75.9 and 72.6\%, so the $+9$-point mean gap includes one failed baseline seed;
Verdict-MQ's win is $+5.2$ and $+7.9$ points against the two converged seeds, and $+2.8$ against the best baseline run
of the whole sweep (77.8\%).

\paragraph{The $k$=0 vs.\ $k$=22 comparison's eligibility and seed provenance.} The pre-registered, dev-split
selection (\texttt{scripts/pick\_split2.py}) that chose the $k$=22 headline recipe considered
$(\text{marker}, k)\in\{\text{mask},\text{cls}\}\times\{0,4,22\}$, a nominal 6 cells, but only kept a cell eligible if
it had a result on all three of RACE, BoolQ and MultiRC; RACE's $k$=4 runs never completed for either marker, so only
4 of the 6 nominal cells (mask/cls $\times$ 0/22) were actually eligible, and $k$=0 was one of those 4 complete,
fully evaluated candidates. The 3-seed test-split comparison reported in Section~\ref{sec:standard} reuses the
original dev-search seed's already-published test accuracy (seed 0: RACE 76.3\%, MultiRC 79.2\%, both already in
Table~\ref{tab:split}) and adds 2 new, previously-untrained seeds (1, 2); it is not three independently fresh seeds.

\paragraph{$k$=0 vs.\ $k$=22 against the matched baseline and Laya.} Against the matched baseline, $k$=0 does
\emph{worse} than $k$=22 on RACE ($-2.7$ to $-2.9$ points vs $k$=22's $-1.2$ to $-1.7$) and more unevenly on MultiRC
($+1.7$ to $+12.8$ points, all significant, vs $k$=22's consistent $+5.2$ to $+13.7$); against zero-shot Laya,
$k$=0 stays far ahead on both benchmarks ($+30$--$31$ RACE, $+23$--$24$ MultiRC), as $k$=22 does.

\section{Full typed-decisions search history}
\label{app:paired-full}
Table~\ref{tab:typed-main-full} extends Table~\ref{tab:typed-main} in the main text with the intermediate marker and
split-depth variants (warm20 recipe, original marker, $k$=22) that led to the headline recipe.

\input{tables/typed_main_full}

Table~\ref{tab:paired-full} extends Table~\ref{tab:paired} in the main text (the 5 headline fixed-marker seeds) with
every other configuration's paired test against Laya, for full transparency about the search that led to the
headline recipe.

\input{tables/typed_paired_full}

\section{Hyper-parameters}
\label{app:hparams}
\begin{table}[h]
\centering\small
\begin{tabular}{@{}lp{0.68\columnwidth}@{}}
\toprule
Setting & Value \\
\midrule
Optimizer & AdamW, weight decay 0.01, gradient clipping 1.0 \\
Schedule & 6\% linear warmup (base), 20\% linear warmup (large, needed for training stability; \S\ref{sec:typed}), linear decay to 0 \\
Learning rate & $3\times10^{-5}$ (base), $2\times10^{-5}$ (large; development-selected) \\
Epochs / batch & 6 epochs; 8 cases (base), 4 cases (large) \\
Context & up to 2,048 tokens per packed case; options capped at 48 tokens \\
Loss & soft cross-entropy + squared EMD on score questions (weight 1) \\
Augmentation & random option subsets for choice questions, $p=0.3$ \\
Precision & bf16 autocast; fp32 master weights \\
Selection & last epoch; recipes chosen on the development split \\
Hardware & one RTX 4090 Laptop GPU (16~GB) \\
\bottomrule
\end{tabular}

\caption{Verdict-MQ training settings.}
\label{tab:hparams}
\end{table}

\paragraph{Truncation scan.} We directly checked whether packing's state-truncation (Section~\ref{sec:model}) ever
triggered on any test case, by comparing each case's state token length against its allotted room at
$\text{max\_len}{=}2048$ across every RACE (1{,}407), BoolQ (2{,}938), MultiRC (344) and typed-decisions (400) test
case (\texttt{scripts/scan\_truncation.py}): zero were truncated, tightest margin 567 spare tokens. A related, sharper
failure mode also exists: \texttt{pack()}'s state-room floor only prevents the state's own budget from going
negative, and does not cap the packed questions/options' combined length, which is never truncated: an un-chunked
MultiRC passage's full candidate-answer set, packed without \texttt{mq\_data.to\_cases()}'s chunking, reaches
6{,}378 tokens against $\text{max\_len}{=}2048$. Every real training/eval path chunks via \texttt{to\_cases()} first
(no reported case exceeds 1{,}481 tokens), but \texttt{pack()} itself did not enforce this, so we added an explicit
check that now raises immediately rather than silently returning an oversized sequence if a caller skips chunking.

\paragraph{bf16 behavioral check, full numbers.} On the final trained checkpoint \texttt{stable-cls-s42} (400 real
typed-decisions test cases, 2{,}000 decisions), checked against the autonomy dial's fitted threshold
($\tau{=}0.520$ at the 90\% target): under bf16, permuting a question's options never changed the picked option or
crossed $\tau$ in any of 400 cases; answering alone instead of packed also never changed the picked option, but did
cross $\tau$ in 3 of 400 cases (0.75\%); comparing bf16 against fp32 directly (same grouping) flipped the argmax on
6 of 2{,}000 decisions (0.3\%) and crossed $\tau$ on 5 (0.25\%).

\section{Large-model recipe selection}
Table~\ref{tab:devgrid} lists the learning-rate, epoch and schedule candidates compared for the 395M model on the
development split (15\% of the benchmark's training cases, held out by case), with the chosen recipe marked.
\input{tables/devgrid}

\section{Split-depth Verdict-MQ}
\label{app:split}
An optional variant lets the state attend to its own question in the top $k$ layers instead of never, by giving each
question its own copy of the lower layers' shared state for those $k$ layers. This trades part of the encode-once
speed-up for question-aware reading; the invariance guarantees hold at every $k$ (checked by \texttt{scripts/test\_mq.py}
at $k$=4 and at full depth). Table~\ref{tab:split} sweeps both option markers
(mask, cls) and $k \in \{0, 4, 22\}$ (full ModernBERT-base depth is 22 layers) on BoolQ, MultiRC and RACE, one seed
each, dev and test accuracy alongside per-passage latency. Depth gives a modest RACE gain (+1.5 points from $k$=0 to
$k$=22) without closing the gap to the matched baseline, and a larger BoolQ gain on the seeds that train normally
(+2.9 points at this seed); MultiRC's win over the baseline does not require depth, since it already holds at $k$=0.
Marker and depth interact rather than acting independently: on BoolQ the cls marker leads at $k$=0 and the mask marker at $k$=4, both are best at $k$=22, and neither
choice was stable across seeds at any $k>0$ (Section~\ref{sec:standard}), so this table reports architecture and
per-passage cost at one seed and makes no stability claim; stability is characterized separately.
\input{tables/split}

\section{Metric validation}
Our implementation of the card's metrics reproduces its input-ignoring Uniform row exactly on KL (0.444), total variation
(0.381) and Brier score (0.238). The card does not state how its prior is fitted or how ECE is binned; our discrete-label
prior gives KL 0.397 against the card's 0.347 and a soft-label prior gives 0.327, so the card's prior lies between the two
constructions. We therefore compare ECE and the prior only among systems we measured ourselves.

\section{Supplementary experiments}
\label{app:supplementary}
Three additional experiments do not bear on the central shared-state accuracy--efficiency tradeoff (Sections
\ref{sec:model}--\ref{sec:standard}) but are reported here in full for completeness.

\subsection{Organizational email decisions}
\label{app:email-full}
To test the decision layer on organizational text rather than a synthetic benchmark, we built three decisions over
public Enron emails \citep{klimt2004enron}: needs-reply (yes/no), priority (score 0--3), and escalation (yes/no),
labelled by two independent LLM teachers with no human annotators (labelling protocol and inter-teacher agreement
below). Fine-tuned Laya, Verdict-MQ, and a plain per-task classifier are close enough on both informative tasks
(93.5\% vs 91.0\% vs 91.3\% on needs-reply; 80.3\% vs 80.7\% vs 79.4\% on priority) that we do not read this as a
clear win for either specialized system, confirming on real email what the typed-decisions results showed; our old
per-question cross-encoder is far behind on both. Zero-shot Laya is close to chance on these decisions, which is the
practical reason an organization needs to specialize its decision layer.

\input{tables/email}

\paragraph{Labelling protocol.} Enron emails were labelled for the three decisions by Qwen3-VL-32B on all 4,000
records (3,625 training, 3,567 distinct texts; 375 validation, 371 distinct texts; zero train/val text overlap) on
each of the three tasks (12,000 task-label rows in total), using the same rubric as an earlier 7B distillation;
DeepSeek~V3.2 labelled just the 375 validation records independently (1,125 task-label rows). The evaluation set of
record is the validation subset where the two agree. Agreement between the two strong teachers was 86.0\% (Cohen's
$\kappa$ 0.69) for needs-reply and 60.9\% (quadratic $\kappa$ 0.61) for priority. For escalation it was 92.5\% but
$\kappa$ was only 0.16, because both teachers marked almost no email as an escalation; we report that task for
completeness only. The original 7B labels disagree with the strong teachers on 14\% of needs-reply and about 46\% of
priority items, which is why we relabelled.

\subsection{Negative result: synthetic-corpus generalists}
\label{app:generalist}
A \emph{generalist} never sees the benchmark's workflows and answers their question schemas zero-shot, the mode in
which the card reports Jev (0.727, matched by our own measurement at 0.734) and Decider (0.768, card only). No open
model is a working generalist on this benchmark: Laya's general checkpoint scores 0.358, below the 0.470 of a prior
that ignores the input (Table~\ref{tab:generalist}).

We built the kind of corpus such models are presumably trained on: Claude Sonnet~5 designed 96 workflows across 24
organizational domains (HR, procurement, logistics, IT operations, legal, compliance, insurance, banking operations,
and others), each with a state schema, six to nine latent factors and five typed questions with criterion
descriptions. Three workflows adjacent to the benchmark's customer-service workflow were removed, and no question
wording matches the benchmark's twenty instructions. For each case we sampled the factors, had DeepSeek~V4.1~Flash or
Gemini~3.1~Flash-Lite render a realistic JSON state, and had three teachers from three model families
(DeepSeek~V4~Pro, Gemini~3~Flash, Qwen~3.6~Plus) each give a distribution as points out of 100; the gold is their
mean. The pilot corpus has 9,277 decisions from 1,859 cases; teachers agree on the argmax 79--83\% of the time,
comparable with the benchmark's own teacher self-agreement of 73.5\%, and only 9\% of gold distributions are near
one-hot.

Verdict-MQ trained on this corpus scores only 0.25--0.34 zero-shot on the benchmark, below the ignore-the-input
prior, regardless of initialization or size (Table~\ref{tab:generalist}); pre-training on it before specialist
fine-tuning gives no gain either (0.679 vs 0.681). The failure is on the student's side, not the teacher's: a 32B
teacher labelling benchmark cases with the corpus prompt agrees 0.699 with the benchmark's gold. Because factors were
sampled uniformly, severe combinations are over-represented, so students trained on this corpus predict high
severity on score questions and systematically avoid the most common benchmark answers on choice questions.
Transferable reading of new rubrics does not emerge from nine thousand synthetic decisions at this model scale, even
on held-out \emph{synthetic} workflows (42--53\%). We report this as a negative result: the specialist results in the
main text do not depend on it, and a working generalist would need far more workflow diversity and realistic
case-distribution priors than we built here.

\input{tables/generalist}

\subsection{Evaluation pitfalls}
\label{app:pitfalls}
Several evaluation errors inflate or distort typed-decision results and are easy to make:
\begin{enumerate}
\item \emph{Stripping option descriptions.} Passing only option keys (``0''--``3'', \texttt{human\_review}) removes the
rubric the benchmark defines as part of the input.
\item \emph{Selecting epochs or recipes on the test split.} Choosing the best epoch on test inflates results; every
reported number uses the last epoch and development-split selection (Section~\ref{sec:setup}).
\item \emph{In-sample thresholds.} Choosing the autonomy threshold on the same decisions it is evaluated on overstates
coverage; Table~\ref{tab:autonomy} uses disjoint halves.
\item \emph{Truncating options away.} Placing options after the input text lets right truncation delete them on long
inputs (about 20\% of our email inputs); options must come first or be protected.
\item \emph{Position shortcuts.} With a fixed option order, a per-question model learns slot positions (0.586 $\to$ 0.349
under permutation); random order at training time or order invariance by construction removes this.
\item \emph{Trivially passing invariance tests.} A model that outputs uniform distributions is perfectly invariant; tests
must also check that representations differ and that the model learns (Section~\ref{sec:model}).
\item \emph{Unsourced baselines and mixed modes.} Leaderboards mix specialist and generalist systems and can carry
unsourced figures; every baseline should be measured, or cited to its source, together with its evaluation mode.
\item \emph{Initialization contamination.} The NLI encoder we use was trained on SuperGLUE BoolQ and MultiRC; on those
benchmarks we start from raw ModernBERT instead (Section~\ref{sec:standard}).
\end{enumerate}

\section{Licences and contamination}
\label{app:licences}
ModernBERT, the tasksource NLI models and Laya are Apache-2.0; typed-decisions is Apache-2.0; BoolQ is CC-BY-SA-3.0,
CLINC150 CC-BY-3.0 and Banking77 CC-BY-4.0; RACE and MultiRC are research-use. The tasksource NLI models list SuperGLUE
BoolQ and MultiRC and GLUE SST-2 among their training tasks (training splits), so on BoolQ and MultiRC our primary runs
start from raw ModernBERT. RACE, CLINC150 and Banking77 are not in that list, and typed-decisions (2026) postdates those
models. Enron-derived labels are released keyed to message hashes, not message bodies. The public-data generalist
(Table~\ref{tab:generalist}) uses up to 3,000 training examples per task from AG News, Emotion, DBpedia-14, Yahoo
Answers Topics, TREC question classification, and TweetEval irony and hate; their terms are recorded in the
repository's \texttt{LICENSES-DATA.md}, and neither source texts nor the trained checkpoint are distributed.

\section{Reproducibility}
Code, scripts and result files are at \coderepo{} (Apache-2.0). \texttt{scripts/make\_tables.py} generates every results table from the result files
(Table~\ref{tab:hparams}, of settings, is written by hand). Per-decision predictions are released for every model in the
paired comparisons (Tables~\ref{tab:paired} and~\ref{tab:standard}), so those tests can be recomputed with
\texttt{scripts/\allowbreak significance.py} (accuracy) and \texttt{scripts/\allowbreak distribution\_significance.py} (KL, Brier, TV);
ablation rows are released as aggregate metrics. The README gives the
commands for the typed-decisions headline; the drivers \texttt{scripts/run\_experiments*.sh} contain the training and
evaluation commands of every run, in the order they were run; \texttt{scripts/test\_mq.py} checks the architecture
guarantees.

\section{Jev measurement record}
\label{app:reproduce-jev}
Jev is not a downloadable checkpoint, so its numbers cannot be reproduced the way Laya's can: they depend on a live,
versioned, vendor-operated service that can change without our knowledge. We call it by the alias
\texttt{typesafe-ai/jev} through Vercel's AI Gateway (\texttt{scripts/eval\_jev\_std.py},
\texttt{scripts/eval\_jev\_typed.py}), which does not expose a pinned semantic version in its response; the closest
record we have is the vendor's own card figure for \texttt{Jev 1.13.0} (Table~\ref{tab:typed-main}), which our
independently measured number closely matches, and the response metadata each call returns (a per-call
\texttt{generationId} and routing/cost fields), which we log but do not print here. All standard-benchmark and
typed-decisions Jev numbers in this paper were measured 2026-09-25 through 2026-09-27; a reader attempting exact
reproduction after that date should treat any discrepancy as a possible vendor-side model update, not necessarily an
error in either measurement.

%% file: tables/matched.tex
\begin{table}[t]
\centering
\small
\begin{adjustbox}{max width=\columnwidth}
\begin{tabular}{lrrrr}
\toprule
Laya encoder init & Acc & KL & Brier & ECE \\
\midrule
Laya arch., seed 0 & 0.774 & 1.857 & 0.281 & 0.181 \\
Laya arch., seed 1 & 0.767 & 1.794 & 0.286 & 0.188 \\
Verdict-MQ, seed 0 & 0.767 & 0.100 & 0.054 & 0.129 \\
Verdict-MQ, seed 1 & 0.758 & 0.108 & 0.060 & 0.130 \\
Verdict-MQ + mixer, seed 0 & 0.550 & 0.262 & 0.147 & 0.051 \\
Verdict-MQ + mixer, seed 1 & 0.769 & 0.100 & 0.054 & 0.134 \\
Released Laya checkpoint & 0.767 & 0.117 & 0.061 & 0.214 \\
\bottomrule
\end{tabular}
\end{adjustbox}
\caption{Matched initialization: both architectures start from Laya's own general-purpose encoder and are trained on the same 6,000 decisions. Our re-fine-tuning of Laya uses hard labels and no temperature, so its distribution metrics are not a fair calibration comparison; the released checkpoint is.}
\label{tab:matched}
\end{table}

%% file: tables/ablations.tex
\begin{table}[t]
\centering
\small
\begin{adjustbox}{max width=\columnwidth}
\begin{tabular}{lrrrr}
\toprule
Variant (one change each) & Acc & KL & Brier & MAE \\
\midrule
Full recipe (3 seeds) & 0.681\,$\pm$\,0.013 & 0.191 & 0.108 & 0.364 \\
no question-type embedding & 0.677 & 0.196 & 0.110 & 0.378 \\
ordered option positions & 0.686 & 0.188 & 0.105 & 0.351 \\
no ordinal (EMD) loss & 0.697 & 0.182 & 0.102 & 0.358 \\
no option-subset augmentation & 0.666 & 0.202 & 0.116 & 0.392 \\
512-token window & 0.671 & 0.193 & 0.106 & 0.341 \\
constant learning rate $2\times10^{-5}$ & 0.733 & 0.161 & 0.083 & 0.307 \\
\bottomrule
\end{tabular}
\end{adjustbox}
\caption{Component ablations of Verdict-MQ on typed-decisions (test split, ModernBERT-base initialization, one seed per variant; the full recipe's seed s.d.\ is 0.013). Only the learning-rate schedule moves accuracy beyond seed noise, and that effect does not carry over to the NLI initializations (Appendix~\ref{sec:schedule}).}
\label{tab:ablations}
\end{table}

%% file: figures/rc.tex
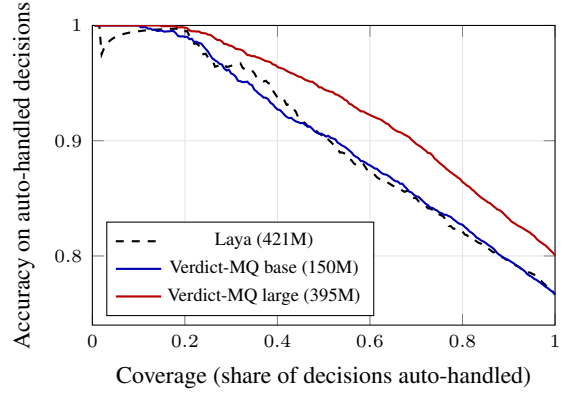
\begin{figure}[t]
\centering
\begin{tikzpicture}
\begin{axis}[width=\columnwidth, height=0.72\columnwidth, xmin=0, xmax=1, ymin=0.74, ymax=1.0,
  xlabel={Coverage (share of decisions auto-handled)}, ylabel={Accuracy on auto-handled decisions},
  legend pos=south west, legend style={font=\scriptsize}, tick label style={font=\scriptsize}, label style={font=\small},
  grid=major, grid style={gray!20}]
\addplot[thick, black, dashed] table[x=coverage, y=accuracy, col sep=comma] {figures/rc_laya.csv};
\addlegendentry{Laya (421M)}
\addplot[thick, blue!70!black] table[x=coverage, y=accuracy, col sep=comma] {figures/rc_mq_base.csv};
\addlegendentry{Verdict-MQ base (150M)}
\addplot[thick, red!70!black] table[x=coverage, y=accuracy, col sep=comma] {figures/rc_mq_large.csv};
\addlegendentry{Verdict-MQ large (395M)}
\end{axis}
\end{tikzpicture}
\caption{Accuracy-coverage curves on the typed-decisions test decisions (decisions ranked by confidence; seeds pooled:
three for the base model, the five fixed-marker seeds 40--44 for the large model; Laya is the released checkpoint).
Higher is better at every coverage. Table~\ref{tab:autonomy} gives the held-out operating points.}
\label{fig:rc}
\end{figure}

%% file: tables/typed_main_full.tex
\begin{table*}[t]
\centering
\small
\begin{adjustbox}{max width=\textwidth}
\begin{tabular}{lrrrrrrrrr}
\toprule
System & Acc & Soft & F1 & KL & TV & Brier & ECE & MAE & W1 \\
\midrule
Uniform (card) & 0.308 & 0.311 & 0.152 & 0.444 & 0.381 & 0.238 & 0.169 & -- & -- \\
Prior (card) & 0.470 & 0.430 & 0.207 & 0.347 & 0.317 & 0.189 & 0.088 & -- & -- \\
MiniLM-L6 per workflow (card) & 0.587 & 0.506 & 0.424 & 0.262 & 0.267 & 0.143 & 0.108 & 0.515 & 0.864 \\
ModernBERT-base per workflow (card) & 0.646 & 0.542 & 0.469 & 0.223 & 0.249 & 0.119 & 0.179 & 0.444 & 0.931 \\
Featherless Simple Jev (card) & 0.716 & -- & -- & 0.488 & -- & 0.176 & -- & -- & -- \\
TypeSafe Jev 1.13.0 (card) & 0.727 & 0.580 & 0.613 & 1.442 & 0.251 & 0.148 & 0.144 & 0.391 & 0.952 \\
meraGPT Decider 1 (card) & 0.768 & 0.608 & 0.641 & 0.096 & 0.149 & 0.052 & 0.180 & 0.219 & 0.984 \\
\midrule
Laya general checkpoint, zero-shot (measured) & 0.358 & 0.347 & 0.239 & 0.570 & 0.410 & 0.316 & 0.178 & 0.695 & 0.726 \\
Laya typed-decisions checkpoint, 421M (measured) & 0.767 & 0.605 & 0.661 & 0.117 & 0.174 & 0.061 & 0.214 & 0.242 & 0.981 \\
TypeSafe Jev, general (measured, Appendix~\ref{app:reproduce-jev}) & 0.734 & 0.583 & 0.639 & 0.829 & 0.251 & 0.148 & 0.028 & 0.391 & 0.966 \\
\midrule
Verdict-CE, per-question cross-encoder, 150M & 0.586 & 0.499 & 0.440 & 0.245 & 0.255 & 0.141 & 0.066 & 0.455 & 0.875 \\
Verdict-MQ, ModernBERT-base init, 3 seeds & 0.681\,$\pm$\,0.013 & 0.553 & 0.545 & 0.191\,$\pm$\,0.009 & 0.213 & 0.108\,$\pm$\,0.006 & 0.095\,$\pm$\,0.010 & 0.364 & 0.934 \\
\quad ModernBERT-large init (raw), 1 seed & 0.729 & 0.575 & 0.608 & 0.153 & 0.184 & 0.089 & 0.126 & 0.309 & 0.955 \\
Verdict-MQ, NLI-base init, 150M, 3 seeds & 0.766\,$\pm$\,0.008 & 0.602 & 0.660 & 0.103\,$\pm$\,0.001 & 0.149 & 0.057\,$\pm$\,0.001 & 0.135\,$\pm$\,0.008 & 0.241 & 0.977 \\
Verdict-MQ, NLI-large init, 395M, 5 fresh seeds & 0.801\,$\pm$\,0.005 & 0.616 & 0.702 & 0.080\,$\pm$\,0.001 & 0.129 & 0.045\,$\pm$\,0.001 & 0.156\,$\pm$\,0.007 & 0.200 & 0.983 \\
\quad fixed marker, $k$=22 (reading-comprehension setting), 5 seeds & 0.804\,$\pm$\,0.003 & 0.617 & 0.710 & 0.079\,$\pm$\,0.001 & 0.128 & 0.044\,$\pm$\,0.001 & 0.160\,$\pm$\,0.004 & 0.199 & 0.986 \\
\quad original marker, warm20 recipe, 5 fresh seeds & 0.790\,$\pm$\,0.007 & 0.612 & 0.682 & 0.087\,$\pm$\,0.002 & 0.134 & 0.048\,$\pm$\,0.001 & 0.148\,$\pm$\,0.006 & 0.210 & 0.979 \\
\quad original marker, earlier recipe (6\% warmup), all 6 seeds & 0.753\,$\pm$\,0.057 & 0.595 & 0.631 & 0.115\,$\pm$\,0.045 & 0.156 & 0.063\,$\pm$\,0.023 & 0.133\,$\pm$\,0.022 & 0.255 & 0.969 \\
\bottomrule
\end{tabular}
\end{adjustbox}
\caption{Full typed-decisions history including intermediate marker and split-depth variants; Table~\ref{tab:typed-main} in the main text gives the headline rows only. Columns as in Table~\ref{tab:typed-main}.}
\label{tab:typed-main-full}
\end{table*}

%% file: tables/typed_paired_full.tex
\begin{table*}[t]
\centering
\small
\begin{adjustbox}{max width=\textwidth}
\begin{tabular}{lrrrrr}
\toprule
Verdict-MQ run & Acc & $\Delta$Acc vs Laya [95\% CI] & item-level McNemar $p$ & case-cluster bootstrap $p$ & $\Delta$ECE [95\% CI] \\
\midrule
NLI-base, seed 0 & 76.3 & $-0.4$ [$-2.0$, $+1.2$] & 0.67 & -- & $-0.082$ [$-0.098$, $-0.066$] \\
NLI-base, seed 1 & 77.5 & $+0.8$ [$-0.8$, $+2.4$] & 0.37 & -- & $-0.070$ [$-0.086$, $-0.053$] \\
NLI-base, seed 2 & 76.1 & $-0.6$ [$-2.2$, $+1.1$] & 0.51 & -- & $-0.085$ [$-0.101$, $-0.068$] \\
NLI-base + option mixer, seed 9 & 77.0 & $+0.2$ [$-1.4$, $+1.9$] & 0.81 & -- & $-0.083$ [$-0.099$, $-0.067$] \\
NLI-base + option mixer, seed 10 & 76.4 & $-0.3$ [$-1.9$, $+1.4$] & 0.77 & -- & $-0.081$ [$-0.098$, $-0.065$] \\
NLI-base + option mixer, seed 11 & 75.9 & $-0.8$ [$-2.3$, $+0.8$] & 0.39 & -- & $-0.087$ [$-0.102$, $-0.070$] \\
NLI-base, constant lr $10^{-5}$, seed 6 & 72.0 & $-4.6$ [$-6.5$, $-2.8$] & 9.8e-07 & -- & $-0.111$ [$-0.129$, $-0.093$] \\
NLI-base, constant lr $10^{-5}$, seed 7 & 73.7 & $-3.0$ [$-4.8$, $-1.2$] & 0.0012 & -- & $-0.109$ [$-0.128$, $-0.091$] \\
NLI-base, constant lr $10^{-5}$, seed 8 & 71.3 & $-5.4$ [$-7.3$, $-3.5$] & 2.5e-08 & -- & $-0.126$ [$-0.144$, $-0.107$] \\
NLI-large, seed 0 & 78.9 & $+2.2$ [$+0.6$, $+3.8$] & 0.0085 & -- & $-0.067$ [$-0.082$, $-0.050$] \\
NLI-large, seed 1 & 70.7 & $-6.0$ [$-7.8$, $-4.2$] & 4.6e-11 & -- & $-0.095$ [$-0.112$, $-0.077$] \\
NLI-large, seed 2 & 65.8 & $-10.9$ [$-13.0$, $-8.7$] & 2.8e-22 & -- & $-0.119$ [$-0.140$, $-0.097$] \\
NLI-large, seed 3 & 79.5 & $+2.8$ [$+1.3$, $+4.4$] & 0.00056 & -- & $-0.060$ [$-0.076$, $-0.044$] \\
NLI-large, seed 4 & 78.5 & $+1.8$ [$+0.3$, $+3.5$] & 0.026 & -- & $-0.068$ [$-0.084$, $-0.052$] \\
NLI-large, seed 5 & 78.3 & $+1.6$ [$+0.1$, $+3.1$] & 0.053 & -- & $-0.077$ [$-0.092$, $-0.060$] \\
NLI-large, lr $10^{-5}$ arm, seed 3 & 77.4 & $+0.7$ [$-0.8$, $+2.3$] & 0.42 & -- & $-0.076$ [$-0.091$, $-0.060$] \\
NLI-large, lr $10^{-5}$ arm, seed 4 & 76.4 & $-0.3$ [$-1.8$, $+1.3$] & 0.75 & -- & $-0.085$ [$-0.100$, $-0.069$] \\
NLI-large, lr $10^{-5}$ arm, seed 5 & 78.5 & $+1.8$ [$+0.3$, $+3.3$] & 0.024 & -- & $-0.071$ [$-0.087$, $-0.056$] \\
NLI-large, original marker, seed 30 & 78.2 & $+1.5$ [$-0.1$, $+3.2$] & 0.078 & -- & $-0.072$ [$-0.087$, $-0.055$] \\
NLI-large, original marker, seed 31 & 79.3 & $+2.6$ [$+1.1$, $+4.2$] & 0.0012 & -- & $-0.060$ [$-0.075$, $-0.044$] \\
NLI-large, original marker, seed 32 & 79.5 & $+2.7$ [$+1.2$, $+4.4$] & 0.0008 & -- & $-0.064$ [$-0.080$, $-0.048$] \\
NLI-large, original marker, seed 33 & 78.2 & $+1.5$ [$-0.0$, $+3.1$] & 0.06 & -- & $-0.073$ [$-0.089$, $-0.057$] \\
NLI-large, original marker, seed 34 & 79.7 & $+2.9$ [$+1.4$, $+4.5$] & 0.00033 & -- & $-0.059$ [$-0.076$, $-0.043$] \\
\textbf{NLI-large, fixed marker, seed 40} & 79.8 & $+3.1$ [$+1.4$, $+4.8$] & 0.00016 & 0.0002 & $-0.059$ [$-0.076$, $-0.041$] \\
\textbf{NLI-large, fixed marker, seed 41} & 80.3 & $+3.6$ [$+1.9$, $+5.4$] & 6.2e-06 & 0.0002 & $-0.054$ [$-0.070$, $-0.035$] \\
\textbf{NLI-large, fixed marker, seed 42} & 80.8 & $+4.0$ [$+2.4$, $+5.7$] & 3.2e-07 & 0.0002 & $-0.048$ [$-0.064$, $-0.031$] \\
\textbf{NLI-large, fixed marker, seed 43} & 80.0 & $+3.2$ [$+1.6$, $+4.9$] & 4.2e-05 & 0.0002 & $-0.060$ [$-0.076$, $-0.044$] \\
\textbf{NLI-large, fixed marker, seed 44} & 79.5 & $+2.8$ [$+1.2$, $+4.5$] & 0.00048 & 0.0018 & $-0.067$ [$-0.083$, $-0.049$] \\
NLI-large, fixed marker, $k$=22, seed 30 & 80.4 & $+3.7$ [$+2.2$, $+5.2$] & 3.3e-06 & -- & $-0.052$ [$-0.066$, $-0.035$] \\
NLI-large, fixed marker, $k$=22, seed 31 & 79.8 & $+3.1$ [$+1.6$, $+4.7$] & 5.9e-05 & -- & $-0.061$ [$-0.076$, $-0.045$] \\
NLI-large, fixed marker, $k$=22, seed 32 & 80.6 & $+3.9$ [$+2.3$, $+5.5$] & 2.4e-06 & -- & $-0.056$ [$-0.072$, $-0.039$] \\
NLI-large, fixed marker, $k$=22, seed 33 & 80.7 & $+4.0$ [$+2.4$, $+5.5$] & 4.7e-07 & -- & $-0.052$ [$-0.067$, $-0.036$] \\
NLI-large, fixed marker, $k$=22, seed 34 & 80.5 & $+3.8$ [$+2.2$, $+5.4$] & 2.4e-06 & -- & $-0.052$ [$-0.067$, $-0.036$] \\
\bottomrule
\end{tabular}
\end{adjustbox}
\caption{Full paired-comparison search history (Table~\ref{tab:paired} in the main text shows only the 5 headline seeds): every seed of every configuration tried against the released Laya typed-decisions checkpoint on the identical 2,000 test decisions (400 cases $\times$ 5 questions each). Item-level exact McNemar treats every decision as independent and is anti-conservative here, since the 5 questions of one case are correlated; the case-cluster bootstrap (10,000 resamples of whole cases) is the primary test. Includes the two large-model seeds that stalled early in training.}
\label{tab:paired-full}
\end{table*}

%% file: tables/devgrid.tex
\begin{table}[t]
\centering
\small
\begin{adjustbox}{max width=\columnwidth}
\begin{tabular}{llrr}
\toprule
Config & Arguments & Dev acc per seed & Mean \\
\midrule
lr2e5e6 (chosen) & {-}{-}lr 2e-5 {-}{-}epochs 6 & 0.792, 0.805 & 0.798 $\pm$ 0.009 \\
lr1e5e6 & {-}{-}lr 1e-5 {-}{-}epochs 6 & 0.793, 0.782 & 0.788 $\pm$ 0.008 \\
lr1e5e10 & {-}{-}lr 1e-5 {-}{-}epochs 10 & 0.787, 0.792 & 0.789 $\pm$ 0.003 \\
lr5e6e10 & {-}{-}lr 5e-6 {-}{-}epochs 10 & 0.786, 0.791 & 0.788 $\pm$ 0.003 \\
const2e5e6 & {-}{-}lr 2e-5 {-}{-}epochs 6 {-}{-}schedule constant & 0.801, 0.757 & 0.779 $\pm$ 0.031 \\
const1e5e6 & {-}{-}lr 1e-5 {-}{-}epochs 6 {-}{-}schedule constant & 0.795, 0.773 & 0.784 $\pm$ 0.016 \\
\bottomrule
\end{tabular}
\end{adjustbox}
\caption{Large-model recipe selection on the development split (15\% of the benchmark's training cases, held out by case). The recipe with the highest mean development accuracy is used for the final test runs.}
\label{tab:devgrid}
\end{table}

%% file: tables/split.tex
\begin{table*}[t]
\centering
\small
\begin{adjustbox}{max width=\textwidth}
\begin{tabular}{lrrrrrrrrr}
\toprule
config & \multicolumn{3}{c}{BoolQ} & \multicolumn{3}{c}{MultiRC} & \multicolumn{3}{c}{RACE} \\ & dev & test & ms & dev & test & ms & dev & test & ms \\
\midrule
mask, $k$=0 & 64.8 & 66.3 & 8.3 & 77.2 & 77.9 & 20.2 & 76.6 & 76.1 & 12.0 \\
mask, $k$=4 & 78.2 & 82.3 & 8.5 & 77.4 & 77.1 & 23.9 & -- & -- & -- \\
mask, $k$=22 & 79.7 & 82.7 & 8.2 & 79.2 & 79.2 & 50.7 & 78.8 & 77.3 & 16.0 \\
cls, $k$=0 & 74.1 & 79.1 & 8.5 & 79.3 & 79.2 & 20.6 & 76.6 & 76.3 & 12.1 \\
cls, $k$=4 & 66.0 & 68.9 & 8.9 & 79.1 & 78.6 & 23.2 & -- & -- & -- \\
cls, $k$=22 (chosen) & 79.3 & 82.0 & 8.2 & 81.0 & 80.1 & 50.9 & 78.9 & 77.8 & 17.2 \\
\midrule tuned baseline & 78.6 & 81.4 & 7.2 & 68.4 & 66.4 & 44.0 & -- & 79.0 & 49.6 \\
\bottomrule
\end{tabular}
\end{adjustbox}
\caption{Marker (mask/cls, Section~\ref{sec:model}) $\times$ split-depth ($k$ = top layers where the state attends to each question; $k$=22 is full ModernBERT-base depth) grid, seed 0. dev is the selection split, test the reporting split; ms = latency per passage (all its questions), batch 1, bf16, RTX 4090 Laptop. RACE's $k$=4 runs never completed for either marker (Appendix~\ref{app:standard-methods}). Baselines: pair classifier (BoolQ, MultiRC; learning rate tuned on dev) and multiple-choice head (RACE). (mask, $k$=4) comes within about a point of the chosen (cls, $k$=22) on BoolQ dev but was not stable across seeds (Section~\ref{sec:standard}); this table reports architecture and cost, not stability.}
\label{tab:split}
\end{table*}

%% file: tables/email.tex
\begin{table}[t]
\centering
\small
\begin{adjustbox}{max width=\columnwidth}
\begin{tabular}{lrrr}
\toprule
Model & Needs reply & Priority (0--3) & Escalate$^\dagger$ \\
\midrule
Majority class & 68.7 & 48.2 & 99.1 \\
Laya, zero-shot & 53.9 & 26.3 & 89.9 \\
Laya, fine-tuned (one model) & 93.5 & 80.3 & 95.1 \\
Plain classifier (one per task) & 91.3 & 79.4 & 96.3 \\
Verdict-CE (per question) & 78.3 & 55.3 & 98.8 \\
Verdict-MQ (NLI-base, one model) & 91.0 & 80.7 & 96.8 \\
\bottomrule
\end{tabular}
\end{adjustbox}
\caption{Organizational email decisions (Enron): accuracy on the consensus validation items where the two strong LLM teachers agree (323, 228 and 347 items). $^\dagger$Escalate is uninformative: only 3 of 347 consensus items are escalations, so the majority class scores 99.1\%.}
\label{tab:email}
\end{table}

%% file: tables/generalist.tex
\begin{table}[t]
\centering
\small
\begin{adjustbox}{max width=\columnwidth}
\begin{tabular}{lrrr}
\toprule
Generalist (zero-shot) & Acc & KL & Brier \\
\midrule
Laya general ckpt (test) & 0.358 & 0.570 & 0.316 \\
Verdict-MQ, public data & 0.359 & 0.560 & 0.309 \\
Verdict-MQ, synthetic corpus & 0.322 & 0.513 & 0.280 \\
\quad + NLI-base init & 0.246 & 0.573 & 0.299 \\
\quad + NLI-large init & 0.338 & 0.541 & 0.305 \\
Prior (card, test) & 0.470 & 0.347 & 0.189 \\
32B teacher, zero-shot & 0.699 & -- & -- \\
\bottomrule
\end{tabular}
\end{adjustbox}
\caption{Negative result: generalists trained on public data or on our synthetic multi-workflow corpus, scored zero-shot on typed-decisions (training split used as the development split, since a generalist never trains on it). All fall below the input-ignoring prior, although the teacher that labelled the corpus agrees 0.699 with the benchmark's gold.}
\label{tab:generalist}
\end{table}

%% file: verdict.bbl
\begin{thebibliography}{46}
\providecommand{\natexlab}[1]{#1}

\bibitem[{Brier(1950)}]{brier1950}
Glenn~W. Brier. 1950.
\newblock Verification of forecasts expressed in terms of probability.
\newblock \emph{Monthly Weather Review}, 78(1):1--3.

\bibitem[{Cao et~al.(2020)Cao, Trivedi, Balasubramanian, and
  Balasubramanian}]{cao2020deformer}
Qingqing Cao, Harsh Trivedi, Aruna Balasubramanian, and Niranjan
  Balasubramanian. 2020.
\newblock {DeFormer}: Decomposing pre-trained transformers for faster question
  answering.
\newblock In \emph{Proceedings of ACL}.

\bibitem[{Cheng et~al.(2023)Cheng, Kasai, and Yu}]{cheng2023batch}
Zhoujun Cheng, Jungo Kasai, and Tao Yu. 2023.
\newblock Batch prompting: Efficient inference with large language model
  {API}s.
\newblock In \emph{Proceedings of the 2023 Conference on Empirical Methods in
  Natural Language Processing: Industry Track}.

\bibitem[{Clark et~al.(2019)Clark, Lee, Chang, Kwiatkowski, Collins, and
  Toutanova}]{clark2019boolq}
Christopher Clark, Kenton Lee, Ming-Wei Chang, Tom Kwiatkowski, Michael
  Collins, and Kristina Toutanova. 2019.
\newblock {BoolQ}: Exploring the surprising difficulty of natural yes/no
  questions.
\newblock In \emph{Proceedings of NAACL-HLT}.

\bibitem[{{Convai Innovations}(2026)}]{laya2026}
{Convai Innovations}. 2026.
\newblock Laya: open typed-decision models.
\newblock Model card, \url{https://huggingface.co/convaiinnovations/laya}.
  Apache-2.0. Vendor source, not peer-reviewed.

\bibitem[{Davison and Hinkley(1997)}]{davison1997bootstrap}
A.~C. Davison and D.~V. Hinkley. 1997.
\newblock \emph{Bootstrap Methods and their Application}.
\newblock Cambridge University Press.

\bibitem[{Deka(2026)}]{deka2026sbert2s1}
Pritam Deka. 2026.
\newblock From retrieval to typed decisions: Calibrated system one models from
  biomedical sentence encoders.
\newblock \emph{arXiv preprint arXiv:2610.02486}.

\bibitem[{Devlin et~al.(2019)Devlin, Chang, Lee, and
  Toutanova}]{devlin2019bert}
Jacob Devlin, Ming-Wei Chang, Kenton Lee, and Kristina Toutanova. 2019.
\newblock {BERT}: Pre-training of deep bidirectional transformers for language
  understanding.
\newblock In \emph{Proceedings of NAACL-HLT}.

\bibitem[{Dodge et~al.(2020)Dodge, Ilharco, Schwartz, Farhadi, Hajishirzi, and
  Smith}]{dodge2020finetuning}
Jesse Dodge, Gabriel Ilharco, Roy Schwartz, Ali Farhadi, Hannaneh Hajishirzi,
  and Noah Smith. 2020.
\newblock Fine-tuning pretrained language models: Weight initializations, data
  orders, and early stopping.
\newblock \emph{arXiv preprint arXiv:2002.06305}.

\bibitem[{Efron and Tibshirani(1993)}]{efron1993bootstrap}
Bradley Efron and Robert~J. Tibshirani. 1993.
\newblock \emph{An Introduction to the Bootstrap}.
\newblock Chapman \& Hall.

\bibitem[{Egressy and St{\"u}hmer(2025)}]{egressy2025setllm}
Beni Egressy and Jan St{\"u}hmer. 2025.
\newblock {Set-LLM}: A permutation-invariant {LLM}.
\newblock In \emph{Advances in Neural Information Processing Systems
  (NeurIPS)}.

\bibitem[{Geifman and El-Yaniv(2017)}]{geifman2017selective}
Yonatan Geifman and Ran El-Yaniv. 2017.
\newblock Selective classification for deep neural networks.
\newblock In \emph{Advances in Neural Information Processing Systems}.

\bibitem[{Gim et~al.(2024)Gim, Chen, Lee, Sarda, Khandelwal, and
  Zhong}]{gim2024promptcache}
In~Gim, Guojun Chen, Seung-seob Lee, Nikhil Sarda, Anurag Khandelwal, and Lin
  Zhong. 2024.
\newblock Prompt cache: Modular attention reuse for low-latency inference.
\newblock In \emph{Proceedings of Machine Learning and Systems (MLSys)}.

\bibitem[{Guo et~al.(2017)Guo, Pleiss, Sun, and
  Weinberger}]{guo2017calibration}
Chuan Guo, Geoff Pleiss, Yu~Sun, and Kilian~Q. Weinberger. 2017.
\newblock On calibration of modern neural networks.
\newblock In \emph{Proceedings of ICML}.

\bibitem[{Hinton et~al.(2015)Hinton, Vinyals, and Dean}]{hinton2015distilling}
Geoffrey Hinton, Oriol Vinyals, and Jeff Dean. 2015.
\newblock Distilling the knowledge in a neural network.
\newblock \emph{arXiv preprint arXiv:1503.02531}.

\bibitem[{Hou et~al.(2016)Hou, Yu, and Samaras}]{hou2016emd}
Le~Hou, Chen-Ping Yu, and Dimitris Samaras. 2016.
\newblock Squared earth mover's distance-based loss for training deep neural
  networks.
\newblock \emph{arXiv preprint arXiv:1611.05916}.

\bibitem[{Humeau et~al.(2020)Humeau, Shuster, Lachaux, and
  Weston}]{humeau2020poly}
Samuel Humeau, Kurt Shuster, Marie-Anne Lachaux, and Jason Weston. 2020.
\newblock Poly-encoders: Architectures and pre-training strategies for fast and
  accurate multi-sentence scoring.
\newblock In \emph{Proceedings of ICLR}.

\bibitem[{Jin et~al.(2026)Jin, Ma, Yin, Chen, Song, Pang, and
  Zhang}]{jin2026duplexjev}
Jie Jin, Ziyin Ma, Min Yin, Jinyu Chen, Haigang Song, Zhikun Pang, and Xiaowen
  Zhang. 2026.
\newblock Batched speech decisions without decoding: Single-token supervision
  lets a frozen {LLM} hear beyond the transcript.
\newblock \emph{arXiv preprint arXiv:2610.02638}.

\bibitem[{Kahneman(2011)}]{kahneman2011thinking}
Daniel Kahneman. 2011.
\newblock \emph{Thinking, Fast and Slow}.
\newblock Farrar, Straus and Giroux.

\bibitem[{Khashabi et~al.(2018)Khashabi, Chaturvedi, Roth, Upadhyay, and
  Roth}]{khashabi2018multirc}
Daniel Khashabi, Snigdha Chaturvedi, Michael Roth, Shyam Upadhyay, and Dan
  Roth. 2018.
\newblock Looking beyond the surface: A challenge set for reading comprehension
  over multiple sentences.
\newblock In \emph{Proceedings of NAACL-HLT}.

\bibitem[{Khattab and Zaharia(2020)}]{khattab2020colbert}
Omar Khattab and Matei Zaharia. 2020.
\newblock {ColBERT}: Efficient and effective passage search via contextualized
  late interaction over {BERT}.
\newblock In \emph{Proceedings of SIGIR}.

\bibitem[{Klimt and Yang(2004)}]{klimt2004enron}
Bryan Klimt and Yiming Yang. 2004.
\newblock The {E}nron corpus: A new dataset for email classification research.
\newblock In \emph{Proceedings of ECML}.

\bibitem[{Krell et~al.(2021)Krell, Kosec, Perez, and
  Fitzgibbon}]{krell2021packing}
Mario~Michael Krell, Matej Kosec, Sergio~P. Perez, and Andrew Fitzgibbon. 2021.
\newblock Efficient sequence packing without cross-contamination: Accelerating
  large language models without impacting performance.
\newblock \emph{arXiv preprint arXiv:2107.02027}.

\bibitem[{Lai et~al.(2017)Lai, Xie, Liu, Yang, and Hovy}]{lai2017race}
Guokun Lai, Qizhe Xie, Hanxiao Liu, Yiming Yang, and Eduard Hovy. 2017.
\newblock {RACE}: Large-scale {R}e{A}ding comprehension dataset from
  examinations.
\newblock In \emph{Proceedings of EMNLP}.

\bibitem[{Lee et~al.(2019)Lee, Lee, Kim, Kosiorek, Choi, and
  Teh}]{lee2019settransformer}
Juho Lee, Yoonho Lee, Jungtaek Kim, Adam Kosiorek, Seungjin Choi, and Yee~Whye
  Teh. 2019.
\newblock Set transformer: A framework for attention-based
  permutation-invariant neural networks.
\newblock In \emph{Proceedings of ICML}.

\bibitem[{Lin(2026)}]{lin2026pact}
Yida Lin. 2026.
\newblock {PACT}: Pairwise-anchored calibrated tuning for single-token typed
  decisions.
\newblock \emph{arXiv preprint arXiv:2609.35865}.

\bibitem[{Liu et~al.(2019)Liu, Ott, Goyal, Du, Joshi, Chen, Levy, Lewis,
  Zettlemoyer, and Stoyanov}]{liu2019roberta}
Yinhan Liu, Myle Ott, Naman Goyal, Jingfei Du, Mandar Joshi, Danqi Chen, Omer
  Levy, Mike Lewis, Luke Zettlemoyer, and Veselin Stoyanov. 2019.
\newblock {RoBERTa}: A robustly optimized {BERT} pretraining approach.
\newblock \emph{arXiv preprint arXiv:1907.11692}.

\bibitem[{{LocalLLaMA}(2026)}]{typeddecisions2026}
{LocalLLaMA}. 2026.
\newblock Typed decisions: a benchmark for typed probabilistic decisions over
  shared state.
\newblock Hugging Face dataset card,
  \url{https://huggingface.co/datasets/LocalLLaMA/typed-decisions}. Not
  peer-reviewed; leaderboard figures other than those we measured are
  card-reported.

\bibitem[{MacAvaney et~al.(2020)MacAvaney, Nardini, Perego, Tonellotto,
  Goharian, and Frieder}]{macavaney2020prettr}
Sean MacAvaney, Franco~Maria Nardini, Raffaele Perego, Nicola Tonellotto, Nazli
  Goharian, and Ophir Frieder. 2020.
\newblock Efficient document re-ranking for transformers by precomputing term
  representations.
\newblock In \emph{Proceedings of SIGIR}.

\bibitem[{McNemar(1947)}]{mcnemar1947}
Quinn McNemar. 1947.
\newblock Note on the sampling error of the difference between correlated
  proportions or percentages.
\newblock \emph{Psychometrika}, 12(2):153--157.

\bibitem[{{meraGPT}(2026)}]{decider2026}
{meraGPT}. 2026.
\newblock Decider 1 (sd-1).
\newblock Proprietary model; figures as reported on the Typed Decisions card.
  Not peer-reviewed.

\bibitem[{Mosbach et~al.(2021)Mosbach, Andriushchenko, and
  Klakow}]{mosbach2021stability}
Marius Mosbach, Maksym Andriushchenko, and Dietrich Klakow. 2021.
\newblock On the stability of fine-tuning {BERT}: Misconceptions, explanations,
  and strong baselines.
\newblock In \emph{Proceedings of ICLR}.

\bibitem[{Naeini et~al.(2015)Naeini, Cooper, and Hauskrecht}]{naeini2015ece}
Mahdi~Pakdaman Naeini, Gregory~F. Cooper, and Milos Hauskrecht. 2015.
\newblock Obtaining well calibrated probabilities using {B}ayesian binning.
\newblock In \emph{Proceedings of AAAI}.

\bibitem[{Peterson et~al.(2019)Peterson, Battleday, Griffiths, and
  Russakovsky}]{peterson2019human}
Joshua~C. Peterson, Ruairidh~M. Battleday, Thomas~L. Griffiths, and Olga
  Russakovsky. 2019.
\newblock Human uncertainty makes classification more robust.
\newblock In \emph{Proceedings of ICCV}.

\bibitem[{Pezeshkpour and Hruschka(2023)}]{pezeshkpour2023order}
Pouya Pezeshkpour and Estevam Hruschka. 2023.
\newblock Large language models sensitivity to the order of options in
  multiple-choice questions.
\newblock \emph{arXiv preprint arXiv:2308.11483}.

\bibitem[{Phang et~al.(2018)Phang, F{\'e}vry, and Bowman}]{phang2018stilts}
Jason Phang, Thibault F{\'e}vry, and Samuel~R. Bowman. 2018.
\newblock Sentence encoders on {STILTs}: Supplementary training on intermediate
  labeled-data tasks.
\newblock \emph{arXiv preprint arXiv:1811.01088}.

\bibitem[{Pruksachatkun et~al.(2020)Pruksachatkun, Phang, Liu, Htut, Zhang,
  Pang, Vania, Kann, and Bowman}]{pruksachatkun2020intermediate}
Yada Pruksachatkun, Jason Phang, Haokun Liu, Phu~Mon Htut, Xiaoyi Zhang,
  Richard~Yuanzhe Pang, Clara Vania, Katharina Kann, and Samuel~R. Bowman.
  2020.
\newblock Intermediate-task transfer learning with pretrained language models:
  When and why does it work?
\newblock In \emph{Proceedings of ACL}.

\bibitem[{Sileo(2024)}]{sileo2024tasksource}
Damien Sileo. 2024.
\newblock tasksource: A large collection of {NLP} tasks with a structured
  dataset preprocessing framework.
\newblock In \emph{Proceedings of LREC-COLING}.

\bibitem[{{TypeSafe AI}(2026)}]{jev2026}
{TypeSafe AI}. 2026.
\newblock Jev: type-safe probabilistic decisions.
\newblock Closed API (\texttt{typesafe-ai/jev} via Vercel AI Gateway). Not
  peer-reviewed; the benchmark card's figure is the card maintainer's own
  measurement through the TypeSafe API.

\bibitem[{Warner et~al.(2024)Warner, Chaffin, Clavi{\'e}, Weller,
  Hallstr{\"o}m, Taghadouini, Gallagher, Biswas, Ladhak, Aarsen, Cooper, Adams,
  Howard, and Poli}]{warner2024modernbert}
Benjamin Warner, Antoine Chaffin, Benjamin Clavi{\'e}, Orion Weller, Oskar
  Hallstr{\"o}m, Said Taghadouini, Alexis Gallagher, Raja Biswas, Faisal
  Ladhak, Tom Aarsen, Nathan Cooper, Griffin Adams, Jeremy Howard, and Iacopo
  Poli. 2024.
\newblock Smarter, better, faster, longer: A modern bidirectional encoder for
  fast, memory efficient, and long context finetuning and inference.
\newblock \emph{arXiv preprint arXiv:2412.13663}.

\bibitem[{Yang et~al.(2022)Yang, Wang, Gan, Zhu, Zhang, Wu, Gao, Zhang, and
  Sakai}]{yang2022unimc}
Ping Yang, Junjie Wang, Ruyi Gan, Xinyu Zhu, Lin Zhang, Ziwei Wu, Xinyu Gao,
  Jiaxing Zhang, and Tetsuya Sakai. 2022.
\newblock Zero-shot learners for natural language understanding via a unified
  multiple choice perspective.
\newblock In \emph{Proceedings of the 2022 Conference on Empirical Methods in
  Natural Language Processing}.

\bibitem[{Yilmaz et~al.(2026)Yilmaz, Tasdemir, and Gozay}]{yilmaz2026lavoir}
Furkan Yilmaz, Habibe~Aleyna Tasdemir, and Muhammed~Faruk Gozay. 2026.
\newblock {LAVOIR}: Teaching a single-pass decision encoder when and what to
  ask with amortized value of information.
\newblock \emph{arXiv preprint arXiv:2609.30706}.

\bibitem[{Yin et~al.(2019)Yin, Hay, and Roth}]{yin2019zeroshot}
Wenpeng Yin, Jamaal Hay, and Dan Roth. 2019.
\newblock Benchmarking zero-shot text classification: Datasets, evaluation and
  entailment approach.
\newblock In \emph{Proceedings of EMNLP-IJCNLP}.

\bibitem[{Zaheer et~al.(2017)Zaheer, Kottur, Ravanbakhsh, Poczos,
  Salakhutdinov, and Smola}]{zaheer2017deepsets}
Manzil Zaheer, Satwik Kottur, Siamak Ravanbakhsh, Barnabas Poczos, Ruslan~R.
  Salakhutdinov, and Alexander~J. Smola. 2017.
\newblock Deep sets.
\newblock In \emph{Advances in Neural Information Processing Systems}.

\bibitem[{Zheng et~al.(2024)Zheng, Zhou, Meng, Zhou, and Huang}]{zheng2024mcq}
Chujie Zheng, Hao Zhou, Fandong Meng, Jie Zhou, and Minlie Huang. 2024.
\newblock Large language models are not robust multiple choice selectors.
\newblock In \emph{Proceedings of ICLR}.

\bibitem[{Zheng et~al.(2023)Zheng, Yin, Xie, Sun, Huang, Yu, Cao, Kozyrakis,
  Stoica, Gonzalez, Barrett, and Sheng}]{zheng2023sglang}
Lianmin Zheng, Liangsheng Yin, Zhiqiang Xie, Chuyue Sun, Jeff Huang, Cody~Hao
  Yu, Shiyi Cao, Christos Kozyrakis, Ion Stoica, Joseph~E. Gonzalez, Clark
  Barrett, and Ying Sheng. 2023.
\newblock {SGLang}: Efficient execution of structured language model programs.
\newblock \emph{arXiv preprint arXiv:2312.07104}.

\end{thebibliography}
